\documentclass{article}

\usepackage{etoolbox}
\newtoggle{arxiv}
\toggletrue{arxiv}

\iftoggle{arxiv}{}{
\PassOptionsToPackage{numbers,sort,compress}{natbib}
\usepackage{neurips_2024}
}

\usepackage[utf8]{inputenc} %
\usepackage[T1]{fontenc}    %
\usepackage{hyperref}       %
\usepackage{url}            %
\usepackage{booktabs}       %
\usepackage{amsfonts}       %
\usepackage{nicefrac}       %
\usepackage{microtype}      %
\usepackage{xcolor}         %
\usepackage{amsmath,amsthm,amssymb}

\usepackage{algorithm}
\usepackage{algorithmic}

\usepackage{caption}
\usepackage{subcaption}
\usepackage{multirow}

\usepackage{xspace}
\usepackage{wrapfig}

\usepackage{pifont}

\usepackage[capitalise]{cleveref}  %

\crefname{section}{\S\hspace{-0.2em}}{\S\S\hspace{-0.2em}}
\crefname{subsection}{\S\hspace{-0.2em}}{\S\S\hspace{-0.2em}}
\crefname{subsubsection}{\S\hspace{-0.2em}}{\S\S\hspace{-0.2em}}

\usepackage{comment}
\usepackage[inline]{enumitem}

\usepackage{graphicx}
\usepackage{grffile}
\usepackage{color}

\usepackage{newtxmath}

\iftoggle{arxiv}{
\usepackage[numbers,sort]{natbib}
}{}

\usepackage{import}

\usepackage{tikz-cd}
\usepackage{tikz}

\newtoggle{todo}
\toggletrue{todo}

\newtoggle{comment}
\togglefalse{comment} %
\newcommand{\TD}[1]{\iftoggle{comment}{\textcolor{magenta}{[TD: #1]}}{}}

\renewcommand{\max}{\mathrm{max}}
\renewcommand{\exp}{\mathrm{exp}}

\newcommand{\diag}{\mathrm{diag}}
\newcommand{\softmax}{\mathrm{softmax}}
\newcommand{\rowmax}{\mathrm{rowmax}}
\newcommand{\rowsum}{\mathrm{rowsum}}
\newcommand{\dsoftmax}{\mathrm{dsoftmax}}

\newcommand{\RR}{\mathbb{R}}

\newcommand{\vQ}{\mathbf{Q}}
\newcommand{\vK}{\mathbf{K}}
\newcommand{\vV}{\mathbf{V}}
\newcommand{\vdQ}{\mathbf{dQ}}
\newcommand{\vdK}{\mathbf{dK}}
\newcommand{\vdV}{\mathbf{dV}}
\newcommand{\vS}{\mathbf{S}}
\newcommand{\vdS}{\mathbf{dS}}
\newcommand{\vP}{\mathbf{P}}
\newcommand{\vdP}{\mathbf{dP}}

\newcommand{\vO}{\mathbf{O}}
\newcommand{\vdO}{\mathbf{dO}}
\newcommand{\vM}{\mathbf{M}}

\newcommand{\fa}{\textsc{FlashAttention}\xspace}
\newcommand{\sysnameone}{\textsc{FlashAttention}\xspace}
\newcommand{\faa}{\textsc{FlashAttention-2}\xspace}
\newcommand{\fat}{\textsc{FlashAttention-3}\xspace}  %

\newtheorem*{theorem*}{Theorem}

\newcommand*\samethanks[1][\value{footnote}]{\footnotemark[#1]}
\usepackage{authblk}
\makeatletter
\renewcommand\AB@affilsepx{ \protect\Affilfont}
\makeatother

\iftoggle{arxiv}{
  \setlength{\textwidth}{6.9in}
  \setlength{\textheight}{9in}
  \setlength{\oddsidemargin}{0in}
  \setlength{\evensidemargin}{0in}
  \setlength{\topmargin}{-0.5in}
  \newlength{\defbaselineskip}
  \setlength{\defbaselineskip}{\baselineskip}
  \setlength{\marginparwidth}{0.8in}
}{
\usepackage[compact]{titlesec}
\titlespacing{\section}{0pt}{*1}{*0}
\titlespacing{\subsection}{0pt}{*1.5}{*0}

\usepackage[subtle, mathdisplays=normal, charwidths=normal, leading=normal]{savetrees}

\addtolength\textfloatsep{-0.5em}
\addtolength\intextsep{-0.2em}

\def\setstretch#1{\renewcommand{\baselinestretch}{#1}}
\setstretch{0.985}
\addtolength{\parskip}{-1pt}

}

\title{FlashAttention-3:\\ Fast and Accurate Attention with Asynchrony and Low-precision}

\iftoggle{arxiv}{
  \author[$^1$]{Jay Shah\thanks{Equal contribution}}
  \author[$^1$]{Ganesh Bikshandi\samethanks}
  \author[$^2$]{Ying Zhang}
  \author[$^{3,4}$]{Vijay Thakkar}
  \author[$^3$]{Pradeep Ramani}
  \author[$^{5,6}$]{Tri Dao}
  \affil[$^1$]{Colfax Research}
  \affil[$^2$]{Meta}
  \affil[$^3$]{NVIDIA}
  \affil[$^4$]{Georgia Tech}
  \affil[$^5$]{Princeton University}
  \affil[$^6$]{Together AI\newline}
  \affil[ ]{\hspace{-2em}{\small\texttt{\{jayhshah,ganesh\}@colfax-intl.com},
      \texttt{yingz@meta.com}, \texttt{\{vithakkar,prraman\}@nvidia.com}, \texttt{tri@tridao.me}}}
}{

\author{%
  Jay Shah\thanks{Equal contribution}\: $^{1}$,
  Ganesh Bikshandi\samethanks\: $^{1}$, Ying Zhang $^{2}$, Vijay Thakkar $^{3,4}$,
  Pradeep Ramani $^{3}$, Tri Dao$^{8,9}$\\
  $^1$ Colfax Research\\
  $^2$ Meta\\
  $^3$ NVIDIA \\
  $^4$ Georgia Institute of Technology\\
  $^8$ Princeton University\\
  $^9$ Together AI\\
  {\small\texttt{\{tri\}@tridao.me}}
}
}

\begin{document}

\maketitle

\begin{abstract}
Attention, as a core layer of the ubiquitous Transformer architecture, is the bottleneck for large language models and long-context applications.
\fa elaborated an approach to speed up attention on GPUs through minimizing memory reads/writes.
However, it has yet to take advantage of new capabilities present in recent
hardware, with \faa achieving only 35\% utilization on the H100 GPU.
We develop three main techniques to speed up attention on Hopper GPUs: exploiting
asynchrony of the Tensor Cores and TMA to (1) overlap overall computation and data movement via warp-specialization and (2) interleave block-wise matmul and softmax operations, and (3)
block quantization and incoherent processing that leverages hardware support for FP8 low-precision.
We demonstrate that our method, \fat, achieves speedup on H100 GPUs
by 1.5-2.0$\times$ with FP16 reaching up to 740 TFLOPs/s (75\% utilization), and with FP8 reaching close to 1.2
PFLOPs/s.
We validate that FP8 \fat achieves 2.6$\times$ lower numerical error than a baseline
FP8 attention.

\end{abstract}

\section{Introduction}
\label{sec:intro}

For the Transformer architecture~\citep{vaswani2017attention}, the attention mechanism constitutes the primary computational bottleneck, since computing the self-attention scores of queries and keys has quadratic scaling in the sequence length.
Scaling attention to longer context will unlock new capabilities (modeling and
reasoning over multiple long
documents~\citep{guo2021longt5,shaham2022scrolls,peng2023yarn} and files in
large codebases~\citep{roziere2023code, li2023starcoder}), new modalities (high-resolution
images~\citep{chen2022scaling}, audio~\citep{gulati2020conformer}, video~\citep{ho2022video}), and new applications (user interaction with long history~\citep{sun2019bert4rec},
agent workflow with long horizon~\citep{yao2022react}).
This has generated significant interest in making attention faster in the long-context regime, including 
by approximation~\citep{katharopoulos2020transformers,choromanski2020rethinking,
tay2020efficient} and software optimization
(\citep{rabe2021self,dao2022flashattention,kwon2023efficient}), or even alternative
architectures~\citep{peng2023rwkv,sun2023retentive,gu2023mamba}.

In this work, we build on the work of \citet{dao2022flashattention} on developing exact-attention algorithms that integrate knowledge of the GPU's execution model and hardware characteristics into their high-level design.
In \citep{dao2022flashattention}, Dao et al. introduced \fa, a novel tiling strategy for parallelizing attention that eliminates intermediate reads/writes to slow global memory through fusing all of the attention operations into a single GPU kernel.
\citet{dao2023flashattention2} restructured the algorithm as \faa to also parallelize over the sequence length dimension and perform the inner loop of the forward pass over blocks of the key and value matrices, thus improving the occupancy and distribution of work on the GPU.
However, we observe that \faa nonetheless achieves poor utilization on newer GPUs relative to optimized matrix-multiplication (GEMM) kernels, such as 35\% vs. 80-90\% on the Hopper H100 GPU.
Partially, this may be attributed to implementation-level differences, such as not using Hopper-specific instructions in place of Ampere ones when targeting the Tensor Cores.
Several work such as ThunkerKitten~\citep{spector2024thunder} and cuDNN 9~\citep{cudnn9} has shown that
with Hopper-specific instructions and tile-based abstractions, one can speedup
attention computation and simplify the implementation.

More fundamentally, \faa's algorithm adheres to a simplified synchronous model and
makes no explicit use of asynchrony and low-precision in its design.
Asynchrony is a result of hardware specialization to accelerate the most important
operations in a ML workload: specific hardware
units performing matrix multiplication (Tensor Cores) or memory loading
(Tensor Memory Accelerator -- TMA), separate from the rest of the CUDA cores performing logic, integer, and floating
point computation.
Low precision such as FP8 in Hopper and FP4 in Blackwell, continuing the trend
of FP16 (Pascal in 2017) and BF16 (Ampere in 2020), is a proven technique to get
double or quadruple throughput for the same power and chip area.
We review the capabilities afforded by Hopper in these directions in \cref{subsec:hardware}.
The technical challenge is to redesign \faa to make use of these hardware
features: asynchrony requires overlapping computation between matmul and softmax
even though one depends on the output of the other, and low-precision requires
care to minimize quantization error, especially in the case of outlier features
in LLMs~\citep{dettmers2208llm, sun2024massive}.

To this end, we propose \fat, which contributes and synthesizes three new ideas to further improve performance on newer GPU architectures:\footnote{We describe our results in the context of NVIDIA's Hopper architecture.
However, our algorithm is operative for any GPU architecture with sufficiently robust asynchronous execution and low-precision capabilities.}

\iftoggle{arxiv}{
\begin{enumerate}
}{
\begin{enumerate}[itemsep=0pt,topsep=0pt,leftmargin=*]
}
\item \textbf{Producer-Consumer asynchrony:} We define a warp-specialized software pipelining scheme that exploits the asynchronous execution of data movement and Tensor Cores by splitting producers and consumers of data into separate warps, thereby extending the algorithm's ability to hide memory and instruction issue latencies.
\item \textbf{Hiding softmax under asynchronous block-wise GEMMs:} We overlap the comparatively low-throughput non-GEMM operations involved in softmax, such as floating point multiply-add and exponential, with the asynchronous WGMMA instructions for GEMM.
As part of this, we rework the \faa algorithm to circumvent certain sequential dependencies between softmax and the GEMMs.
For example, in the 2-stage version of our algorithm, while softmax executes on one block of the scores matrix, WGMMA executes in the asynchronous proxy to compute the next block.
\item \textbf{Hardware-accelerated low-precision GEMM:} We adapt the forward pass algorithm to allow for targeting the FP8 Tensor Cores for GEMM, nearly doubling the measured TFLOPs/s. 
This requires bridging the different layout conformance requirements of WGMMA in terms of how blocks of FP32 accumulator and FP8 operand matrices are assumed to be laid out in memory.
We use the techniques of block quantization and incoherent processing to mitigate the loss of accuracy that results from moving to FP8 precision.
\end{enumerate}

To validate our method empirically, we benchmark \fat on the H100 SXM5 GPU over
a range of parameters and show that (1) FP16 achieves 1.5-2.0$\times$ speedup over
\faa in the forward pass (reaching up to 740 TFLOPs/s) and 1.5-1.75$\times$ in the backward pass,
(2) FP8 achieves close to 1.2 PFLOPs/s, and
(3) for large sequence length, FP16 outperforms and FP8 is
competitive\footnote{More precisely, for head dimension 64 \fat FP8 is ahead,
while for head dimensions 128 and 256 it is at par for those cases
without causal masking and behind with causal masking.}
with a state-of-the-art implementation of attention from NVIDIA's cuDNN library.
We also validate that FP16 \fat yields the same numerical error as \faa and is
better than the standard attention implementation as intermediate results (e.g.,
softmax rescaling) are kept in FP32.
Moreover, FP8 \fat with block quantization and incoherent processing is 2.6$\times$ more accurate than standard
attention with per-tensor quantization in cases with outlier features.

We open-source \fat with a permissive license\footnote{\fat
  is available at \url{https://github.com/Dao-AILab/flash-attention}} and plan
to integrate it with
PyTorch and Hugging Face libraries to benefit the largest number of researchers
and developers.

\section{Background: Multi-Head Attention and GPU Characteristics}
\label{sec:background}

\subsection{Multi-Head Attention}
\label{subsec:multi_head_attn}

Let $\vQ, \vK, \vV \in \mathbb{R}^{N \times d}$ be the query, key and value input sequences associated to a single head, where $N$ is the sequence length and $d$ is the head dimension. Then the attention output $\vO$ is computed as:
\begin{equation*}
  \vS = \alpha \vQ \vK^\top \in \mathbb{R}^{N \times N}, \quad \vP = \softmax(\vS) \in \mathbb{R}^{N \times N}, \quad \vO = \vP\vV \in \mathbb{R}^{N \times d},
\end{equation*}
where $\softmax$ is applied row-wise and one typically sets $\alpha = 1/\sqrt{d}$ as the scaling factor.
In practice, we subtract $\rowmax(\vS)$ from $\vS$ to prevent numerical instability with the exponential function.
For multi-head attention (MHA), each head has its own set of query, key and value projections, and this computation parallelizes across multiple heads and batches to produce the full output tensor.

Now let $\phi$ be a scalar loss function and let $\mathbf{d}(-) = \partial \phi / \partial (-)$ be notation for the gradient.
Given the output gradient $\vdO \in \mathbb{R}^{N \times d}$, we compute $\vdQ$, $\vdK$, and $\vdV$ according to the chain rule as follows:
\iftoggle{arxiv}{
\begin{align*}
  \vdV &= \vP^\top \vdO \in \mathbb{R}^{N \times d} \\
  \vdP &= \vdO \vV^\top \in \mathbb{R}^{N \times N} \\
  \vdS &= \dsoftmax (\vdP) \in \mathbb{R}^{N \times N} \\
  \vdQ &= \alpha \vdS \vK \in \mathbb{R}^{N \times d} \\
  \vdK &= \alpha \vdS^\top \vQ \in \mathbb{R}^{N \times d},
\end{align*}
}{
\begin{align*}
  \vdV &= \vP^\top \vdO \in \mathbb{R}^{N \times d}, \: &\vdP &= \vdO \vV^\top \in \mathbb{R}^{N \times N}, \\
  \vdS &= \dsoftmax (\vdP) \in \mathbb{R}^{N \times N}, \: &\vdQ &= \alpha \vdS \vK \in \mathbb{R}^{N \times d}, \qquad \vdK = \alpha \vdS^\top \vQ \in \mathbb{R}^{N \times d}.
\end{align*}
}
Here, we have that $\mathbf{d}s = (\diag(p) - p p^\top)\mathbf{d}p$ for $p = \softmax(s)$ as a function of a vector $s$, and we write $\dsoftmax(\vdP)$ for this formula applied row-wise.
Finally, this computation again parallelizes across the number of heads and batches for the backward pass of MHA.

\subsection{GPU hardware characteristics and execution model}
\label{subsec:hardware}

We describe the aspects of the GPU's execution model relevant for \fat, with a focus on the NVIDIA Hopper architecture as a concrete instantiation of this model.

\paragraph{Memory hierarchy:} The GPU's memories are organized as a hierarchy of data locales, with capacity inversely related to bandwidth (\cref{tab:gpu-hierarchy})\footnote{\citet{luo2024benchmarking} reports shared memory bandwidth of 128 bytes per clock cycle per SM, and we multiply that by 132 SMs and the boost clock of 1830 MHz.}.
Global memory (GMEM), also known as HBM, is the off-chip DRAM accessible to all streaming multiprocessors (SMs).
Data from GMEM gets transparently cached into an on-chip L2 cache.
Next, each SM contains a small on-chip, programmer-managed highly banked cache called shared memory (SMEM). 
Lastly, there is the register file within each SM.

\paragraph{Thread hierarchy:} The GPU's programming model is organized around logical groupings of execution units called threads.
From the finest to coarsest level, the thread hierarchy is comprised of threads, warps (32 threads), warpgroups (4 contiguous warps), threadblocks (i.e., cooperative thread arrays or CTAs), threadblock clusters (in Hopper), and grids.

These two hierarchies are closely interlinked.
Threads in the same CTA are co-scheduled on the same SM, and CTAs in the same cluster are co-scheduled on the same GPC.
SMEM is directly addressable by all threads within a CTA, whereas each thread has at most 256 registers (RMEM) private to itself.

\begin{table}[h!]
  \small
  \centering
  \caption{Thread-Memory hierarchy for the NVIDIA Hopper H100 SXM5 GPU.}
  \label{tab:gpu-hierarchy}
  \begin{tabular}{|r|l|l|l|}
      \hline
      \textbf{Hardware Level} & \textbf{Parallel Agent} & \textbf{Data Locale} & \textbf{Capacity @ Bandwidth} \\
      \hline
      Chip   & Grid                 & GMEM & 80  GiB @ 3.35 TB/s \\
      GPC    & Threadblock Clusters & L2   & 50  MiB @ ~12 TB/s \\
      SM     & Threadblock (CTA)    & SMEM & 228 KiB per SM, 31TB/s per GPU \\
      Thread & Thread               & RMEM & 256 KiB per SM \\
      \hline
  \end{tabular}
\end{table}

\paragraph{Asynchrony and warp-specialization:}

GPUs are throughput processors that rely on concurrency and asynchrony to hide memory and execution latencies.
For async memory copy between GMEM and SMEM, Hopper has the Tensor Memory Accelerator (TMA) as a dedicated hardware unit \cite[\S7.29]{cuda}.
Furthermore, unlike prior architectures such as Ampere, the Tensor Core of Hopper, exposed via the warpgroup-wide WGMMA instruction \cite[\S9.7.14]{ptx}, is also asynchronous and can source its inputs directly from shared memory.

Hardware support for asynchrony allows for warp-specialized kernels, where the warps of a CTA are divided into producer or consumer roles that only ever issue either data movement or computation.
Generically, this improves the compiler's ability to generate optimal instruction schedules \citep{warp-specialization-2011}.
In addition, Hopper supports the dynamic reallocation of registers between warpgroups via \verb|setmaxnreg| \cite[\S9.7.17.1]{ptx}, so those warps doing MMAs can obtain a larger share of RMEM than those just issuing TMA (for which only a single thread is needed).

\paragraph{Low-precision number formats:}
\label{sec:low-precision-gpu}
Modern GPUs have specialized hardware units for accelerating low-precision computation.
For example, the WGMMA instruction can target the FP8 Tensor Cores on Hopper to deliver 2x the throughput per SM when compared to FP16 or BF16.

However, correctly invoking FP8 WGMMA entails understanding the layout constraints on its operands. 
Given a GEMM call to multiply $A \times B^{\top}$ for an $M\times K$-matrix $A$ and an $N\times K$-matrix $B$, we say that the $A$ or $B$ operand is \emph{mn-major} if it is contiguous in the outer $M$ or $N$ dimension, and \emph{k-major} if is instead contiguous in the inner $K$-dimension.
Then for FP16 WGMMA, both mn-major and k-major input operands are accepted for operands in SMEM, but for FP8 WGMMA, only the k-major format is supported.
Moreover, in situations such as attention where one wants to fuse back-to-back GEMMs in a single kernel, clashing FP32 accumulator and FP8 operand layouts pose an obstacle to invoking dependent FP8 WGMMAs.

In the context of attention, these layout restrictions entail certain modifications to the design of an FP8 algorithm, which we describe in \cref{sec:algofp8}.

\subsection{Standard Attention and Flash Attention}
Following \citet{dao2022flashattention}, we let \textbf{standard attention} denote an implementation of attention on the GPU that materializes the intermediate matrices $\vS$ and $\vP$ to HBM. The main idea of \fa was to leverage a local version of the softmax reduction to avoid these expensive intermediate reads/writes and fuse attention into a single kernel. Local softmax corresponds to lines \ref{code-ws:softmax_start}-\ref{code-ws:softmax_end} of the consumer mainloop in \cref{alg:flash3_wgmma_ws_only} together with the rescalings of blocks of $\vO$. The simple derivation that this procedure indeed computes $\vO$ can be found in \cite[\S 2.3.1]{dao2023flashattention2}.

\section{FlashAttention-3: Algorithm}
\label{sec:algo}

In this section, we describe the \fat algorithm. For simplicity, we focus on the
forward pass, with the backward pass algorithm described in~\cref{sec:algo_ws_bwd}. We first indicate how to integrate warp-specialization with a circular SMEM buffer into the base algorithm of \faa. We then explain how to exploit asynchrony of WGMMA to define an overlapped GEMM-softmax 2-stage pipeline. Finally, we describe the modifications needed for FP8, both in terms of layout conformance and accuracy via block quantization and incoherent processing.

\subsection{Producer-Consumer asynchrony through warp-specialization and
  pingpong scheduling}
\label{sec:algo_ws}

\paragraph{Warp-specialization}
As with \faa, the forward pass of \fat is embarrassingly parallel in the batch size, number of heads, and query sequence length.
Thus, it will suffice to give a CTA-level view of the algorithm, which operates on a tile $\vQ_i$ of the query matrix to compute the corresponding tile $\vO_i$ of the output.
To simplify the description, we first give the warp-specialization scheme with a circular SMEM buffer that does \emph{not} have in addition the GEMM-softmax overlapping.
Let $d$ be the head dimension, $N$ the sequence length, and fix a query block size $B_r$ to divide $\vQ$ into $T_r = \lceil \frac{N}{B_r} \rceil$ blocks $\vQ_1, .., \vQ_{T_r}$.

\begin{algorithm}[H]
    \caption{\small\label{alg:flash3_wgmma_ws_only}\fat forward pass \textbf{without} intra-consumer overlapping -- CTA view}
    \begin{algorithmic}[1]
\REQUIRE Matrices $\vQ_i \in \RR^{B_r \times d}$ and $\vK, \vV \in \mathbb{R}^{N \times d}$ in HBM, key block size $B_c$ with $T_c = \lceil \frac{N}{B_c} \rceil$.
\STATE Initialize pipeline object to manage barrier synchronization with $s$-stage circular SMEM buffer.
\IF {in producer warpgroup}
\STATE Deallocate predetermined number of registers.
\STATE Issue load $\vQ_i$ from HBM to shared memory.
\STATE Upon completion, commit to notify consumer of the load of $\vQ_i$.
\FOR{$0 \le j < T_c$}
    \STATE Wait for the $(j\,\%\,s)$th stage of the buffer to be consumed.
    \STATE Issue loads of $\vK_j, \vV_j$ from HBM to shared memory at the $(j\,\%\,s)$th stage of the buffer.
    \STATE Upon completion, commit to notify consumers of the loads of $\vK_j, \vV_j$.
\ENDFOR
\ELSE
\STATE Reallocate predetermined number of registers as function of number of consumer warps.
\STATE On-chip, initialize $\vO_i = (0) \in \mathbb{R}^{B_r \times d}$ and $\ell_i, m_i = (0), (-\infty) \in \mathbb{R}^{B_r}$.
\STATE Wait for $\vQ_i$ to be loaded in shared memory.
\FOR{$0 \le j < T_c$}
\STATE Wait for $\vK_j$ to be loaded in shared memory.
\STATE Compute $\vS_i^{(j)} = \vQ_i \vK_j^T$ (SS-GEMM). Commit and wait. \label{alg:ws_only_gemm1}
\STATE Store $m_i^{\mathrm{old}} = m_i$ and compute $m_i = \max(m_i^{\mathrm{old}}, \rowmax(\vS_i^{(j)}))$. \label{code-ws:softmax_start}
\STATE Compute $\widetilde{\vP}_i^{(j)} = \exp(\vS_i^{(j)} - m_i)$ and $\ell_i = \exp(m_i^{\mathrm{old}} - m_i) \ell_i + \rowsum(\widetilde{\vP}_i^{(j)})$. \label{code-ws:softmax_end}
\STATE Wait for $\vV_j$ to be loaded in shared memory.
\STATE Compute $\vO_i = \diag(\exp(m_i^{\mathrm{old}} - m_i))^{-1} \vO_i + \widetilde{\vP}_i^{(j)} \vV_j$ (RS-GEMM). Commit and wait. \label{alg:ws_only_gemm2}
\STATE Release the $(j\,\%\,s)$th stage of the buffer for the producer.
\ENDFOR
\STATE Compute $\vO_i = \diag(\ell_i)^{-1} \vO_i$ and $L_i = m_i + \log(\ell_i)$.
\STATE Write $\vO_i$ and $L_i$ to HBM as the $i$th block of $\vO$ and $L$.
\ENDIF
\end{algorithmic}
\end{algorithm}

For our implementation of \cref{alg:flash3_wgmma_ws_only} on Hopper, we use \verb|setmaxnreg| for (de)allocations, TMA for loads of $\vQ_i$ and $\{ \vK_j, \vV_j \}_{0 \leq j < T_c}$, and WGMMA to execute the GEMMs in the consumer mainloop, with the SS or RS prefix indicating whether the first operand is sourced from shared memory or register file.
For interpreting the execution flow of \cref{alg:flash3_wgmma_ws_only}, note that issuing TMA loads does not stall on the completion of other loads due to asynchrony.
Moreover, in the producer mainloop, no waits will be issued for the first $s$ iterations as the buffer gets filled.

\paragraph{Pingpong scheduling}
The asynchronous nature of WGMMA and TMA, along with warp-specialization, opens
up the opportunity to overlap the softmax computation of one warpgroup with the GEMM of
another warpgroup.
To motivate this, notice that non-matmul operations have much lower throughput
than matmul operations on modern hardware accelerators.
As an example, the H100 SXM5 GPU has 989 TFLOPS of FP16 matmul but only 3.9
TFLOPS of special functions such as exponential\footnote{The CUDA programming
  guide specifies that 16 operations of special functions can be performed per
  streaming multiprocessor (SM) per clock cycle. We multiply 16 by 132 SMs and
  1830 MHz clock speed to get 3.9 TFLOPS of special functions.} (necessary for softmax).
For the attention forward pass in FP16 with head dimension 128, there are 512x more matmul FLOPS
compared to exponential operations, but the exponential has 256x lower
throughput, so exponential can take 50\% of the cycle compared to matmul.
The situation is even worse with FP8, where the matmul throughput doubles but
the exponential throughput stays the same.

Since the exponential is performed by a separate hardware unit (the multi-function
unit), ideally we'd want the exponential calculation to be scheduled when the
Tensor Cores are performing the matmul.
To do so, we use synchronization barriers (\texttt{bar.sync} instructions) to
force the GEMMs (GEMM1 -- $\vP \vV$ of one iteration, and GEMM0 -- $\vQ \vK^\top$
of the next iteration) of warpgroup 1 to be scheduled before the GEMMs of
warpgroup 2.
As a result, the softmax of warpgroup 1 will be scheduled while warpgroup 2 is
performing its GEMMs. Then the roles swap, with warpgroup 2 doing softmax while
warpgroup 1 doing GEMMs (hence, ``pingpong'' scheduling).
This is illustrated in~\cref{fig:pingpong_scheduling}.
Though in practice the pingpong scheduling is not as clean as depicted in the
figure, we generally find this to improve performance (e.g., from 570 TFLOPS to
620-640 TFLOPS for FP16 forward with head dimension 128 and sequence length 8192).
\begin{figure}[ht]
    \centering
    \includegraphics[width=1.0\linewidth]{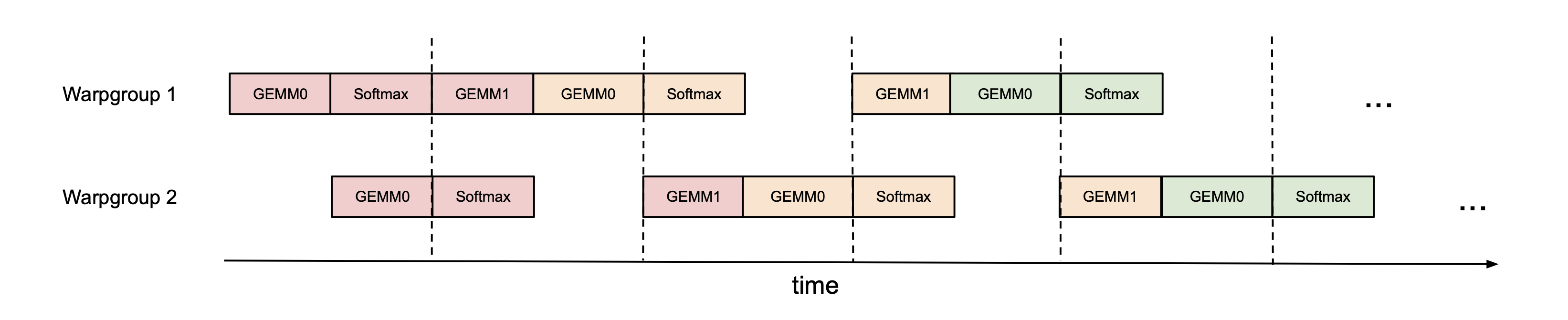}
    \caption{Pingpong scheduling for 2 warpgroups to overlap softmax and GEMMs: the softmax of one warpgroup
      should be scheduled when the GEMMs of another warpgroup are running. The
      same color denotes the same iteration.}
    \label{fig:pingpong_scheduling}
\end{figure}

\paragraph{Attention variants}
For multi-query attention~\citep{shazeer2019fast} and grouped query
attention~\citep{ainslie2023gqa}, we follow the approach in \faa and adjust the
tensor indexing to avoid duplicating $\vK$ and $\vV$ in HBM.

\subsection{Intra-warpgroup overlapping GEMMs and softmax}

Even within one warpgroup, we can overlap some instructions in the softmax with
some instructions in the GEMMs. We describe one technique to do so.

In the attention algorithm, operations within the inner loop (main loop) have sequential dependencies that impede parallelization within a single iteration.
For example, (local) softmax (lines \ref{code-ws:softmax_start} to \ref{code-ws:softmax_end}) relies on the output $\vS_i^{(j)}$ of the first GEMM, while the second GEMM takes its result $\widetilde{\vP}_i^{(j)}$ as an operand.
Indeed, the wait statements in lines \ref{alg:ws_only_gemm1} and \ref{alg:ws_only_gemm2} of \cref{alg:flash3_wgmma_ws_only} serialize the execution of softmax and GEMMs.
However, we can break these dependencies by pipelining across iterations through additional buffers in registers.
Pursuing this idea, we propose the following two-stage\footnote{Note that the number of stages of the overlapping scheme is bounded by, but need not equal, the number $s$ of stages in the circular SMEM buffer.} GEMM-softmax pipelining algorithm:

\begin{figure}[ht]
    \centering
    \includegraphics[width=.95\linewidth]{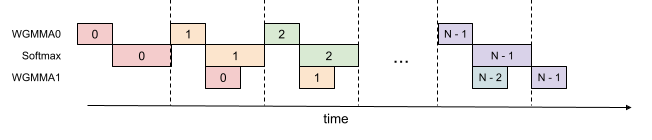}
    \caption{2-stage WGMMA-softmax pipelining}
    \label{fig:2_stage_pipelining}
\end{figure}

\begin{algorithm}[H]
    \caption{\small\label{alg:flash3_wgmma}\fat consumer warpgroup forward pass}
    \begin{algorithmic}[1]
      \REQUIRE  Matrices $\vQ_i \in \RR^{B_r \times d}$ and $\vK, \vV \in \mathbb{R}^{N \times d}$ in HBM, key block size $B_c$ with $T_c = \lceil \frac{N}{B_c} \rceil$.
      \STATE Reallocate predetermined number of registers as function of number of consumer warps.
      \STATE On-chip, initialize $\vO_i = (0) \in \mathbb{R}^{B_r \times d}$ and $\ell_i, m_i = (0), (-\infty) \in \mathbb{R}^{B_r}$.
      \STATE Wait for $\vQ_i$ and $\vK_0$ to be loaded in shared memory.
      \STATE Compute $\vS_{\mathrm{cur}} = \vQ_i \vK_0^T$ using WGMMA. Commit and wait.
      \STATE Release the $0$th stage of the buffer for $\vK$.
      \STATE Compute $m_{i}$, $\tilde{\vP}_{\mathrm{cur}}$ and $\ell_{i}$ based on $\vS_{\mathrm{cur}}$, and rescale $\vO_i$.
      \FOR{$1 \le j < T_c - 1$}
        \STATE Wait for $\vK_j$ to be loaded in shared memory.
        \label{code:mainloop_start}
        \STATE Compute $\vS_{\mathrm{next}} = \vQ_i \vK_{j}^T$ using WGMMA. Commit but do not wait. \label{code:first_wgmma}
        \STATE Wait for $\vV_{j-1}$ to be loaded in shared memory.
        \STATE Compute $\vO_{i} = \vO_{i} + \tilde{\vP}_{\mathrm{cur}} \vV_{j-1}$ using WGMMA. Commit but do not wait. \label{code:second_wgmma}
        \STATE Wait for the WGMMA $\vQ_i \vK_{j}^T$.
        \STATE Compute $m_{i}$, $\tilde{\vP}_{\mathrm{next}}$ and $\ell_{i}$ based on $\vS_{\mathrm{next}}$. \label{code:softmax}
        \STATE Wait for the WGMMA $\tilde{\vP}_{\mathrm{cur}} \vV_{j-1}$ and then rescale $\vO_i$
        \STATE Release the $(j\,\%\,s)$th, resp. $(j-1\,\%\,s)$th stage of the buffer for $\vK$, resp. $\vV$.
        \STATE Copy $\vS_{\mathrm{next}}$ to $\vS_{\mathrm{cur}}$.
        \label{code:mainloop_end}
      \ENDFOR
      \STATE Wait for $\vV_{T_c - 1}$ to be loaded in shared memory.
      \STATE Compute
      $\vO_{i} = \vO_{i} + \tilde{\vP}_{\mathrm{last}} \vV_{T_c - 1}$ using WGMMA. Commit and wait.

      \STATE Epilogue:
      Rescale $\vO_{i}$ based on $m_{i}$.
      Compute $L_{i}$ based on $m_{i}$ and $\ell_{i}$.
      Write $\vO_{i}$ and $L_{i}$ to HBM as the $i$-th block of $\vO$ and $L$.
    \end{algorithmic}
  \end{algorithm}

  \cref{alg:flash3_wgmma} functions as a replacement for the consumer path of \cref{alg:flash3_wgmma_ws_only} to comprise the complete \fat algorithm for FP16 precision. At a high-level, we use WGMMA as a metonym for asynchronous GEMM. Within the mainloop (lines \ref{code:mainloop_start} to \ref{code:mainloop_end}), the second WGMMA operation of iteration $j$ (line \ref{code:second_wgmma}) is overlapped with softmax operations from iteration $j+1$ (line \ref{code:softmax}).

  While the pipelined structure illustrated above offers theoretical performance gains, there are several practical aspects to consider:
  \paragraph{Compiler reordering}
  The pseudocode represents an idealized execution order but the compiler (NVCC) often rearranges instructions for optimization.
  This can disrupt the carefully crafted WGMMA and non-WGMMA operation pipelining sequence, potentially leading to unexpected behavior or diminished performance gains. An analysis of the SASS code shows that the compiler generates overlapped code as expected (Section~\ref{sec:2-stage-sass}).
  \paragraph{Register pressure}
  To maintain optimal performance, register spilling should be minimized.
  However, the 2-stage pipeline requires additional registers to store intermediate results and maintain context between stages.
  Specifically, an extra $\vS_{\mathrm{next}}$ must be kept in registers, leading to extra register usage of size $B_r \times B_c \times \text{sizeof}(\text{float})$ per threadblock.
  This increased register demand may conflict with using larger block sizes (another common optimization), which is also register-hungry.
  In practice, trade-offs should be made based on profiling results.
  \paragraph{3-stage pipelining} Extending the 2-stage algorithm described above, we propose a 3-stage variant
  that would further overlap the second WGMMA with softmax.
  While this approach offers the potential for even higher Tensor Core utilization,
  it requires even more registers due to an additional stage in the pipeline,
  making the trade-off between tile size and pipeline depth more difficult to balance.
  A detailed description of the 3-stage algorithm and its evaluation results can be found in ~\cref{sec:3-stage}.

\subsection{Low-precision with FP8}
\label{sec:algofp8}

\begin{figure} %
\centering
\begin{tikzpicture}
\definecolor{mygold}{RGB}{255, 215, 0} %
\definecolor{warmcolor}{RGB}{255, 165, 0} %
\def\r{.9}
\def\h{-.5}
\foreach \y in {0,2} {
  \ifnum\y=0
    \def\mycolor{cyan!20}
  \else
    \def\mycolor{mygold!20}
  \fi
  \foreach \x in {0,2} {
    \draw[fill=\mycolor] (4*\y*\r+\x*\r,0) rectangle ++(2*\r,.5);
    \draw[fill=\mycolor] (4*\r+4*\y*\r+\x*\r,0) rectangle ++(2*\r,.5);
  }
}
\foreach \y in {0,2} {
  \ifnum\y=0
    \def\mycolor{cyan!50}
  \else
    \def\mycolor{warmcolor!50}
  \fi
  \foreach \x in {0,2} {
    \draw[fill=\mycolor] (4*\y*\r+\x*\r,\h) rectangle ++(2*\r,.5);
    \draw[fill=\mycolor] (4*\r+4*\y*\r+\x*\r,\h) rectangle ++(2*\r,.5);
  }
}

\foreach \y in {0,2} {
  \foreach \x in {0} {
    \ifnum\y=0
      \node at (4*\y*\r+\x*\r + \r, 0.25) {T0\,\{d0,\,d1\}};
    \else
      \node at (4*\y*\r+\x*\r + \r, 0.25) {T0\,\{d4,\,d5\}};
    \fi
  }
  \foreach \x in {2} {
    \ifnum\y=0
      \node at (4*\y*\r+\x*\r + \r, 0.25) {T1\,\{d0,\,d1\}};
    \else
      \node at (4*\y*\r+\x*\r + \r, 0.25) {T1\,\{d4,\,d5\}};
    \fi
  }
}
\foreach \y in {1,3} {
  \foreach \x in {0} {
  \ifnum\y=1
    \node at (4*\y*\r+\x*\r + \r, 0.25) {T2\,\{d0,\,d1\}};
  \else
    \node at (4*\y*\r+\x*\r + \r, 0.25) {T2\,\{d4,\,d5\}};
  \fi
  }
  \foreach \x in {2} {
  \ifnum\y=1
    \node at (4*\y*\r+\x*\r + \r, 0.25) {T3\,\{d0,\,d1\}};
  \else
    \node at (4*\y*\r+\x*\r + \r, 0.25) {T3\,\{d4,\,d5\}};
  \fi
  }
}

\foreach \y in {0,2} {
    \foreach \x in {0} {
      \ifnum\y=0
        \node at (4*\y*\r+\x*\r + \r, 0.25+\h) {T0\,\{d2,\,d3\}};
      \else
        \node at (4*\y*\r+\x*\r + \r, 0.25+\h) {T0\,\{d6,\,d7\}};
      \fi
    }
    \foreach \x in {2} {
      \ifnum\y=0
        \node at (4*\y*\r+\x*\r + \r, 0.25+\h) {T1\,\{d2,\,d3\}};
      \else
        \node at (4*\y*\r+\x*\r + \r, 0.25+\h) {T1\,\{d6,\,d7\}};
      \fi
    }
  }

\foreach \y in {1,3} {
  \foreach \x in {0} {
  \ifnum\y=1
    \node at (4*\y*\r+\x*\r + \r, 0.25+\h) {T2\,\{d2,\,d3\}};
  \else
    \node at (4*\y*\r+\x*\r + \r, 0.25+\h) {T2\,\{d6,\,d7\}};
  \fi
  }
  \foreach \x in {2} {
  \ifnum\y=1
    \node at (4*\y*\r+\x*\r + \r, 0.25+\h) {T3\,\{d2,\,d3\}};
  \else
    \node at (4*\y*\r+\x*\r + \r, 0.25+\h) {T3\,\{d6,\,d7\}};
  \fi
  }
}
\end{tikzpicture}
\caption{FP32 accumulator register WGMMA layout -- rows 0 and 8, threads 0-3, entries 0-7.}
\label{fig:rmem_accum}
\end{figure}

\begin{figure}
\centering
\begin{tikzpicture}
\definecolor{mygold}{RGB}{255, 215, 0} %
\definecolor{warmcolor}{RGB}{255, 165, 0} %
\def\r{.9}
\def\h{-.5}

\foreach \y in {0,...,3} {
  \foreach \x in {0,2} {
    \ifnum\x=0
      \def\mycolor{cyan!20}
    \else
      \def\mycolor{cyan!50}
    \fi
    \draw[fill=\mycolor] (4*\y*\r+\x*\r,0) rectangle ++(2*\r,.5);
  }
}
\foreach \y in {0,...,3} {
  \foreach \x in {0,2} {
    \ifnum\x=0
      \def\mycolor{mygold!20}
    \else
      \def\mycolor{warmcolor!50}
    \fi
    \draw[fill=\mycolor] (4*\y*\r+\x*\r,\h) rectangle ++(2*\r,.5);
  }
}
\foreach \y in {0,...,3} {
    \foreach \x in {0,2} {
      \ifnum\x=0
        \node at (4*\y*\r+\x*\r + \r, 0.25) {T\y\,\{a0,\,a1\}};
      \else
        \node at (4*\y*\r+\x*\r + \r, 0.25) {T\y\,\{a2,\,a3\}};
      \fi
    }
}
\foreach \y in {0,...,3} {
    \foreach \x in {0,2} {
      \ifnum\x=0
        \node at (4*\y*\r+\x*\r + \r, 0.25+\h) {T\y\,\{a4,\,a5\}};
      \else
        \node at (4*\y*\r+\x*\r + \r, 0.25+\h) {T\y\,\{a6,\,a7\}};
      \fi
    }
}
\end{tikzpicture}
\caption{FP8 operand A register WGMMA layout -- rows 0 and 8, threads 0-3, entries 0-7.}
\label{fig:rmem_operand}
\end{figure}

\textbf{Efficiency: layout transformations.}
Computing the forward pass of \fat in FP8 precision poses additional challenges not encountered for FP16 in terms of layout conformance.

First, we note that the input tensors $\vQ$, $\vK$, and $\vV$
are typically given as contiguous in the head dimension,
while to satisfy the k-major constraint on FP8 WGMMA for the second GEMM we need $\vV$,
or rather the tiles of $\vV$ loaded into SMEM, to be contiguous in the sequence length dimension.
Since the TMA load itself cannot change the contiguous dimension, we then need to either
(1) transpose $\vV$ in GMEM as a pre-processing step, or
(2) do an in-kernel transpose of tiles of $\vV$ after loading them into SMEM.
To implement option (1), we can either
(1a) fuse the transpose to the epilogue of a preceding step such as the rotary embedding, or
(1b) call a standalone pre-processing transpose
kernel\footnote{An optimized transpose kernel will achieve speed near the bandwidth of the device~\citep{colfax_cutlass_transpose_2024}.}
to exchange the strides of the sequence length and head dimensions.
However, (1a) is difficult to integrate into a standard library,
and (1b) is too wasteful in a memory-bound situation such as inference.

Instead, for FP8 \fat we opt for option (2).
For the in-kernel transpose, we take advantage of the LDSM (\verb|ldmatrix|)
and STSM (\verb|stmatrix|) instructions,
which involve a warp of threads collectively loading SMEM to RMEM
and storing RMEM to SMEM at a granularity of 128 bytes.\footnote{In the PTX documentation, LDSM/STSM are described as copying $8 \times 8$ matrices with 16-bit entries \cite[\S 9.7.13.4.15-16]{ptx},
but we can pack 8-bit entries two at a time to use LDSM/STSM in the context of FP8 precision.
However, the transpose versions of LDSM/STSM cannot split packed 8-bit entries,
which necessitates certain register movements in between LDSM and STSM to actually perform a tile-wise transpose; we omit the details.}
The LDSM/STSM instructions are both register efficient,
allowing us to execute them in the producer warpgroup,
and capable of transposing layouts when doing memory copy.
Moreover, after the first iteration we can arrange
for the transpose of the next $\vV$ tile
to be executed in the shadow of the two WGMMAs
that involve the preceding $\vV$ and current $\vK$ tile.

Second, we observe that unlike with FP16,
the memory layout of the FP32 accumulator of an FP8 WGMMA is different
from that assumed for its operand A when held in registers.
We depict fragments of these two layouts in \cref{fig:rmem_accum} and \cref{fig:rmem_operand},
where the entries are held in registers per thread in the listed order.
By using byte permute instructions,
we can then transform the first WGMMA's accumulator into a format suitable for the second WGMMA,
and compatibly with the layout of the $\vV$ tile produced by the in-kernel transpose. Specifically, with reference to \cref{fig:rmem_accum}, we change the order in sequence to
$$\{ \verb|d0 d1 d4 d5 d2 d3 d6 d7| \},$$
and this register permutation is then replicated over every 8 bytes. In terms of the logical shape of the $\vP$ tile, this manuever permutes its columns (e.g., columns $0189$ now become the first four columns). For WGMMA to then compute the correct output tile, we can correspondingly arrange for the in-kernel transpose to write out a matching row permutation of the $\vV$ tile.\footnote{This additional freedom afforded by doing the in-kernel transpose eliminates having to use shuffle instructions to change register ownership across threads, which we previously described in~\citep{colfax_fp8_flashattention_2024}.}

\textbf{Accuracy: block quantization and incoherent processing.}
With FP8 (e4m3) format, one only uses 3 bits to store the mantissa and 4 bits
for the exponent.
This results in higher numerical error than FP16/BF16.
Moreover, large models typically have outlier values~\citep{dettmers2208llm,
  sun2024massive} that are much larger in magnitude than most other values,
making quantization difficult.
One typically use per-tensor scaling~\citep{micikevicius2022fp8} by keeping one scalar per tensor (e.g., one
for $\vQ$, for $\vK$, and for $\vV$).
To reduce the numerical error of attention in FP8, we employ two techniques:
\iftoggle{arxiv}{
\begin{enumerate}
}{
\begin{enumerate}[itemsep=0pt,topsep=0pt,leftmargin=*]
}
\item \textbf{Block quantization}: we keep one scalar per block, so that
 for each of $\vQ$, $\vK$, $\vV$ we split the tensor into blocks of
  size $B_r \times d$ or $B_c \times d$ and quantize them separately.
  This quantization can be fused with an operation right before attention (e.g.,
  rotary embedding) with no additional slow down (since rotary embedding is
  memory-bandwidth bound).
  As the \fat algorithm naturally operates on blocks, we can scale each block of
  $\vS$ to account for this block quantization at no computation cost.
\item \textbf{Incoherent processing}: to even out outliers, we multiply
  $\vQ$ and $\vK$ with a random orthogonal matrix $\vM$ before quantizing to
  FP8. Since $\vM$ is orthogonal, $\vM \vM^\top = I$ and so $(\vQ \vM) (\vK
  \vM)^\top = \vQ \vK^\top$, i.e., multiplying both $\vQ$ and $\vK$ with
  $\vM$ does not change the attention output.
  This serves to ``spread out'' the outliers since each entry of $\vQ \vM$
  or $\vK \vM$ is a random sum of entries of $\vQ$ or $\vK$, thus reducing
  quantization error.
  In practice, we follow \citet{chee2024quip} and~\citet{tseng2024quip} and choose $\vM$ to be the product of random diagonal matrices of $\pm
  1$ and a Hadamard matrix, which can be multiplied in $O(d \log d)$ instead of
  $O(d^2)$, and can also be fused with the rotary embedding at no extra computation cost.
\end{enumerate}
We validate that these two techniques reduces numerical error by up to 2.6$\times$ in \cref{sec:numerical_error}.

\section{Empirical Validation}
\label{sec:exp}

We use the primitives from CUTLASS~\citep{Thakkar_CUTLASS_2023} such as WGMMA
and TMA abstractions to implement \fat and evaluate its efficiency and accuracy.
\iftoggle{arxiv}{
\begin{itemize}
}{
\begin{itemize}[itemsep=3pt,topsep=3pt,leftmargin=*]
}
\item \textbf{Benchmarking attention.}
We measure the runtime of \fat across different sequence lengths and
compare it to a standard implementation in PyTorch,
\faa, \faa in Triton (which uses H100-specific instructions),
as well as a vendor's implementation of \faa optimized for H100 GPUs
from cuDNN.
We confirm that \fat is up to 2.0$\times$ faster than \faa and 1.5$\times$
faster than \faa in Triton.
\fat reaches up to 740 TFLOPs/s, 75\% of the theoretical maximum TFLOPs/s
on H100 GPUs.
\item \textbf{Ablation study.} We confirm that our algorithmic improvements with warp-specialization and GEMM-softmax pipelining contribute to the speedup of \fat.
\item \textbf{Accuracy of FP8 attention.} We validate that block quantization and incoherent processing reduces the numerical error of FP8 \fat by 2.6$\times$.
\end{itemize}

\subsection{Benchmarking Attention}
\label{subsec:benchmark_attn}

We measure the runtime of different attention methods on an H100 80GB SXM5 GPU
for different settings (without / with causal mask, head dimension 64 or 128) for FP16 inputs.
We report the results
in~\cref{fig:benchmark_attn_fwd} and~\cref{fig:benchmark_attn_bwd}, showing that \fat is around 1.5-2.0$\times$ faster
than \faa in the forward pass and 1.5-1.75$\times$ faster in
the backward pass.
Compared to a standard attention implementation, \fat can be up
to 3-16$\times$ faster.
For medium and long sequences (1k and above), \fat even surpasses the speed of a vendor's library (cuDNN -- closed source) that has been optimized for H100 GPUs.

\paragraph{Benchmark settings:} We vary the sequence length as 512, 1k, ..., 16k, and set
batch size so that the total number of tokens is 16k.
We set the hidden dimension to 2048, and head dimension to be either 64, 128, or 256
(i.e., 32 heads, 16 heads, or 8 heads).
To calculate the FLOPs of the forward pass, we use:
\begin{equation*}
  4 \cdot \text{seqlen}^2 \cdot \text{head dimension} \cdot \text{number of heads}.
\end{equation*}
With causal masking, we divide this number by 2 to account for the fact that
approximately only half of the entries are calculated.
To get the FLOPs of the backward pass, we multiply the forward pass FLOPs by 2.5
(since there are 2 matmuls in the forward pass and 5 matmuls in the backward
pass, due to recomputation).

\begin{figure}[!ht]
  \centering
  \begin{subfigure}{.5\textwidth}
    \centering
    \includegraphics[width=.95\linewidth]{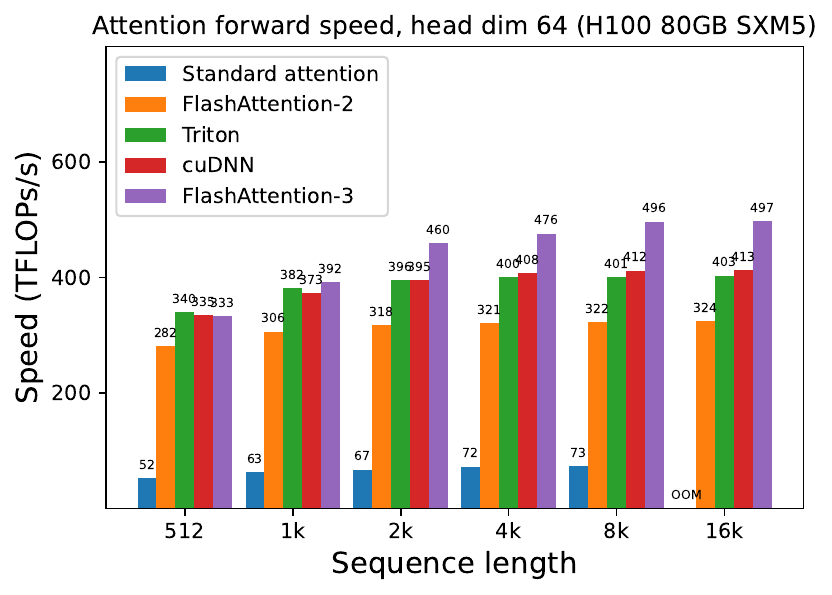}
    \caption{Forward, without causal mask, head dim 64}
  \end{subfigure}%
  \begin{subfigure}{.5\textwidth}
    \centering
    \includegraphics[width=.95\linewidth]{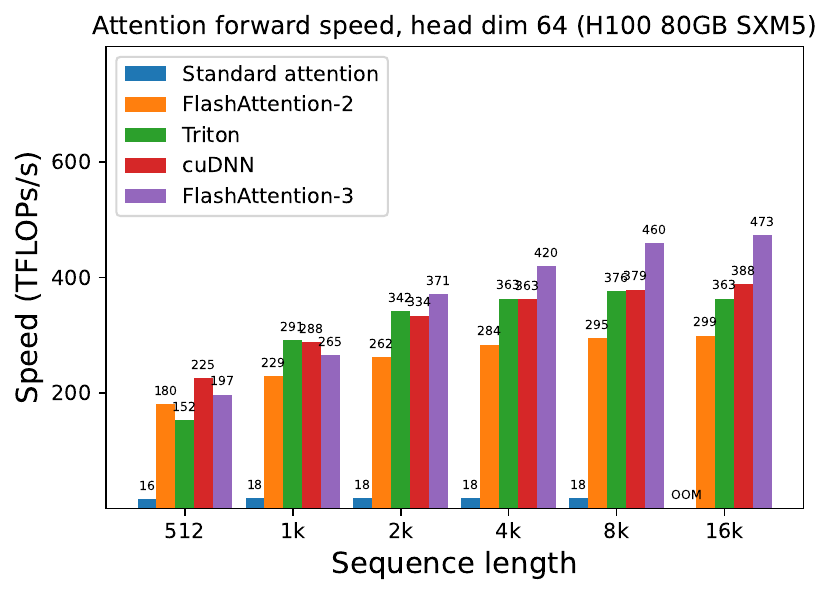}
    \caption{Forward, with causal mask, head dim 64}
  \end{subfigure}
  \begin{subfigure}{.5\textwidth}
    \centering
    \includegraphics[width=.95\linewidth]{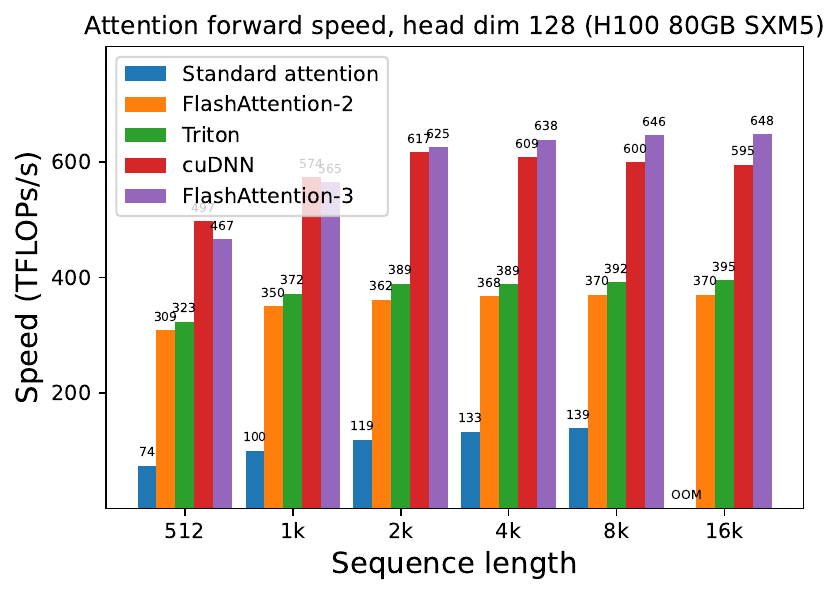}
    \caption{Forward, without causal mask, head dim 128}
  \end{subfigure}%
  \begin{subfigure}{.5\textwidth}
    \centering
    \includegraphics[width=.95\linewidth]{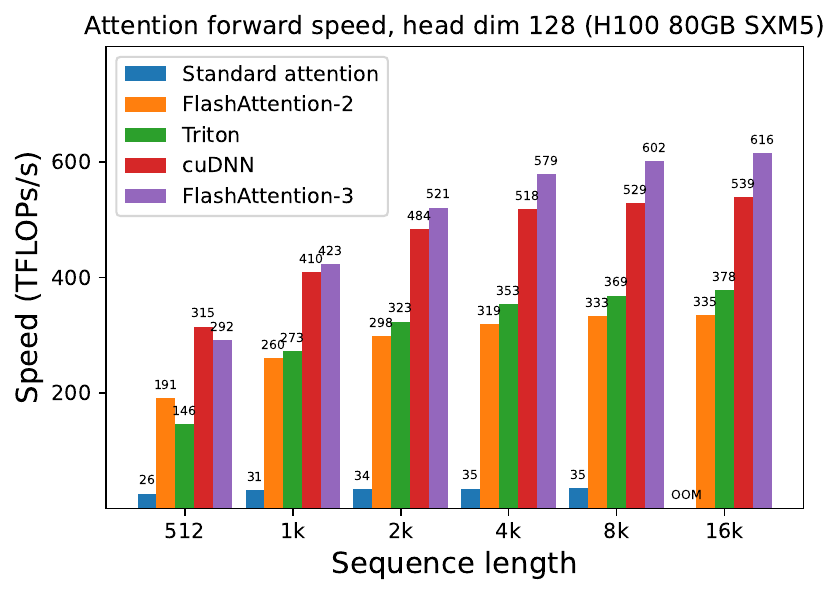}
    \caption{Forward, with causal mask, head dim 128}
  \end{subfigure}
  \begin{subfigure}{.5\textwidth}
    \centering
    \includegraphics[width=.95\linewidth]{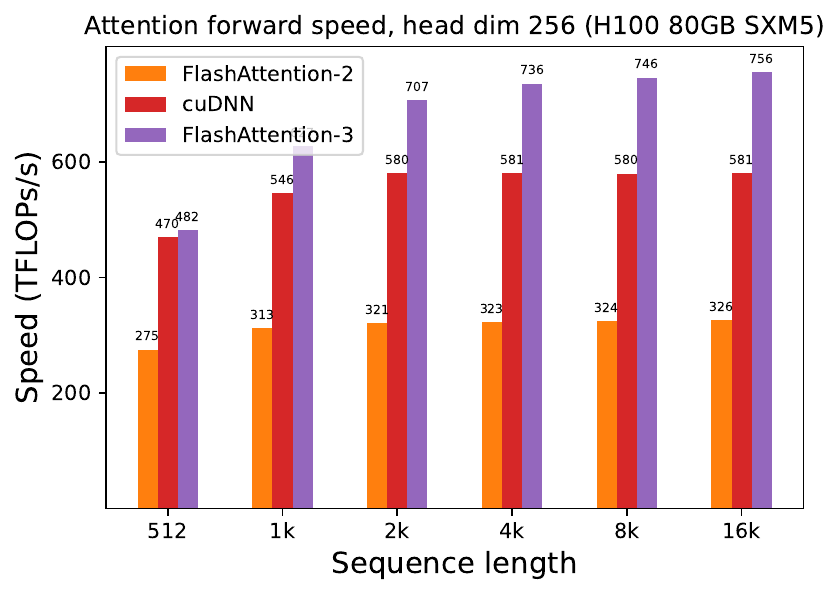}
    \caption{Forward, without causal mask, head dim 256}
  \end{subfigure}%
  \begin{subfigure}{.5\textwidth}
    \centering
    \includegraphics[width=.95\linewidth]{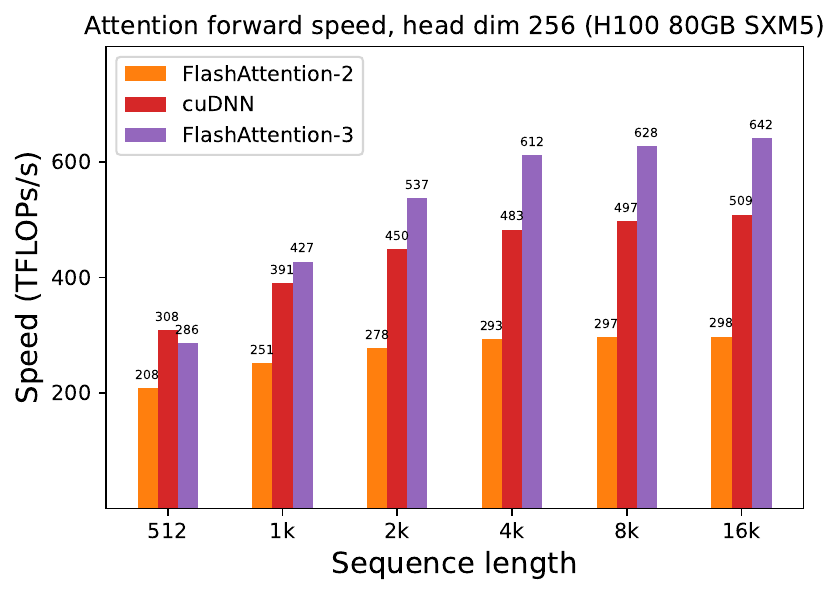}
    \caption{Forward, with causal mask, head dim 256}
  \end{subfigure}
  \caption{Attention forward speed (FP16/BF16) on H100 GPU}
  \label{fig:benchmark_attn_fwd}
\end{figure}

\begin{figure}[ht]
  \centering
  \begin{subfigure}{.5\textwidth}
    \centering
    \includegraphics[width=.95\linewidth]{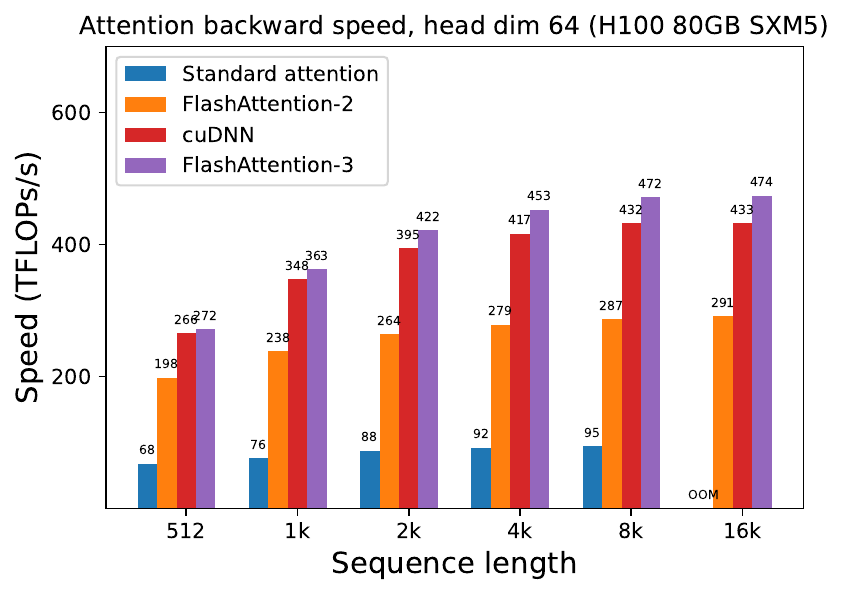}
    \caption{Backward, without causal mask, head dim 64}
  \end{subfigure}%
  \begin{subfigure}{.5\textwidth}
    \centering
    \includegraphics[width=.95\linewidth]{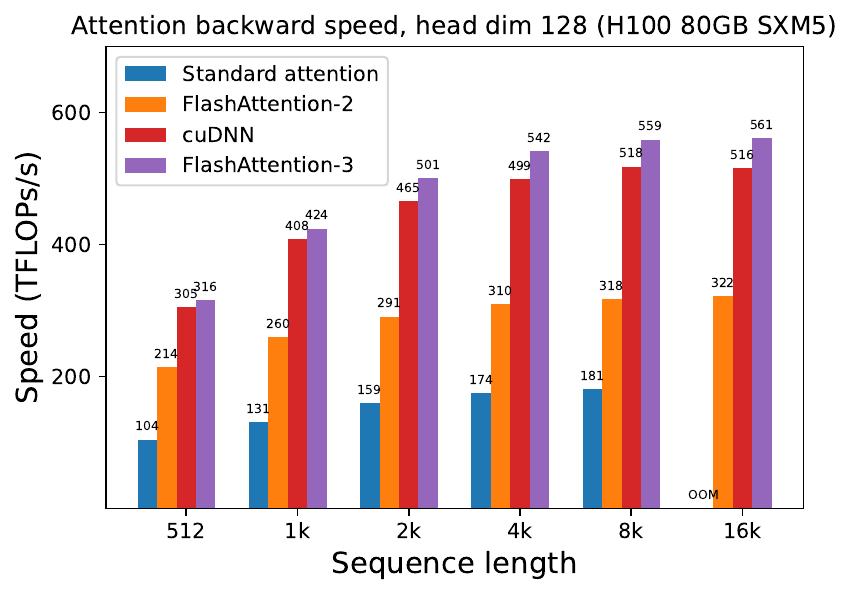}
    \caption{Backward, without causal mask, head dim 128}
  \end{subfigure}
  \caption{Attention backward speed (FP16/BF16) on H100 GPU}
  \label{fig:benchmark_attn_bwd}

\end{figure}

We also measure the runtime for FP8 for the forward pass under similar settings.
We report the results for headdim 256 in ~\cref{fig:benchmark_attn_fp8} and give the full results in \cref{sec:benchmark_attn_fp8_full}.

\begin{figure}[ht]
  \centering
  \begin{subfigure}{.5\textwidth}
    \centering
    \includegraphics[width=.95\linewidth]{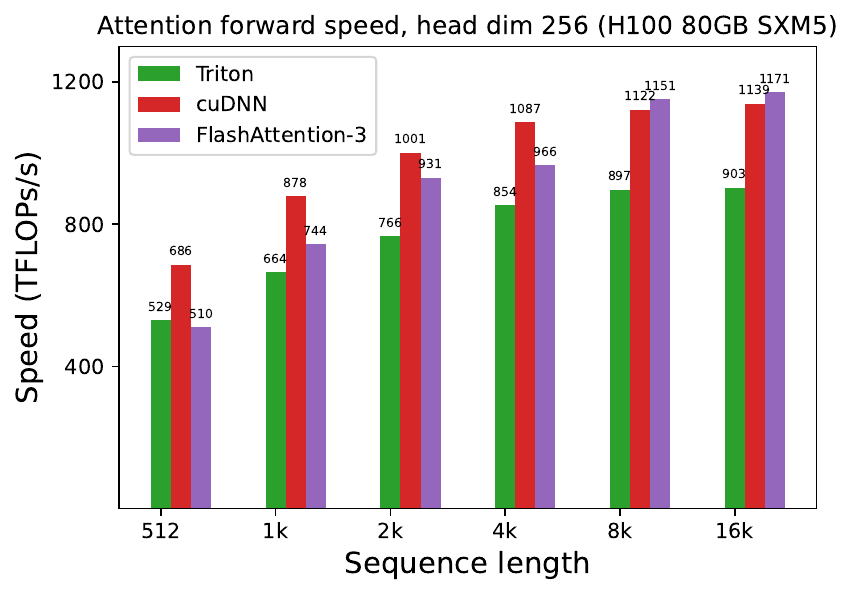}
    \caption{Forward, without causal mask, head dim 256}
  \end{subfigure}%
  \begin{subfigure}{.5\textwidth}
    \centering
    \includegraphics[width=.95\linewidth]{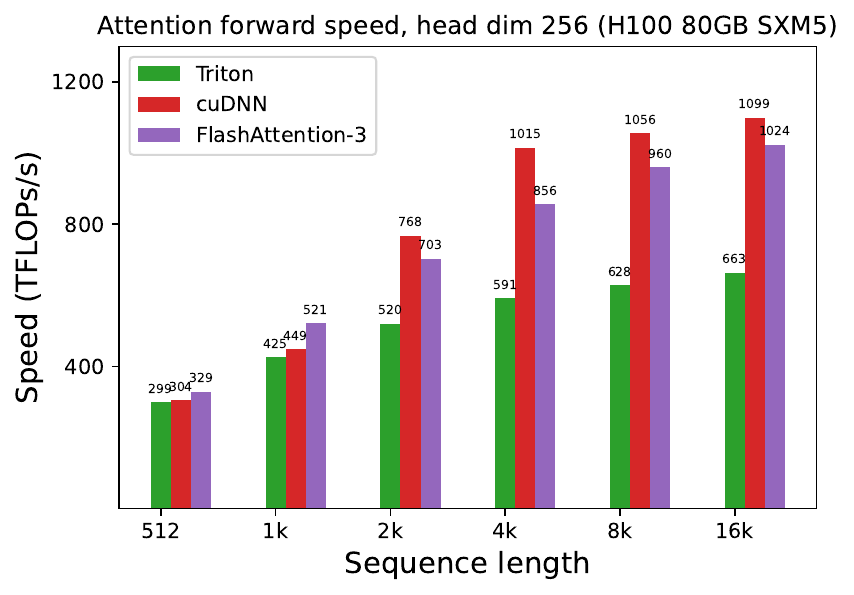}
    \caption{Forward, with causal mask, head dim 256}
  \end{subfigure}
  \caption{Attention forward speed (FP8) on H100 GPU}
  \label{fig:benchmark_attn_fp8}
\end{figure}

\subsection{Ablation Study: 2-Stage Pipelining Experiments}
\label{sec:2-stage-pipelining-experiments}

We ablate both the 2-stage WGMMA-softmax pipelining and warp-specialization for non-causal FP16 \fat with fixed parameters $\{\text{batch}, \text{seqlen}, \text{nheads}, \text{hdim}\} = \{ 4, 8448, 16, 128\}$.
The result in~\cref{table:ablation_pipelining} confirms that our algorithmic improvements (asynchrony with warp-specialization and overlapping between GEMM and softmax) lead to significant speedup, from 570 to 661 TFLOPs.
\begin{table}[h!]
  \centering  
  \caption{Pipelining ablation measurements}
  \label{table:ablation_pipelining}
  \begin{tabular}{|l|l|l|}
      \hline
      \textbf{Configuration} & \textbf{Time} & \textbf{TFLOPs/s} \\
      \hline
      \fat & 3.538 ms & 661 \\
      \hline
      No GEMM-Softmax Pipelining, Warp-Specialization & 4.021 ms & 582 \\
      \hline
      GEMM-Softmax Pipelining, No Warp-Specialization & 4.105 ms & 570 \\
      \hline
  \end{tabular}

\end{table}

\subsection{Numerical Error Validation}
\label{sec:numerical_error}

As there has been interest in the numerical error~\citep{golden2024flash} of
\fa, we compare \faa, \fat, and a standard implementation of attention against a
reference implementation in FP64.
To simulate outlier features and activations in LLMs~\citep{dettmers2208llm,
  sun2024massive}, we generate the entries of $\vQ, \vK, \vV$ with the following
distribution:
\begin{equation*}
  \mathcal{N}(0, 1) + \mathcal{N}(0, 100) \cdot \mathrm{Bernoulli}(0.001).
\end{equation*}
That is, each entry is normally distributed with zero mean and standard
deviation 1, but for 0.1\% of entries we add an independent term that's normally distributed
with standard deviation 10.
We then measure the root mean squared error (RMSE) in~\cref{table:numerical_error}.
In FP16, both \faa and \fat achieves 1.7$\times$ lower RMSE compared to the standard implementation since intermediate results (softmax) are kept in FP32.
The baseline attention in FP8 uses per-tensor scaling, with matmul accumulator in FP32 and intermediate softmax results kept in FP16.
Thanks to block quantization and incoherent processing, \fat in FP8 is 2.6$\times$ more accurate than this baseline.
\begin{table}[h!]
  \centering
  \caption{Numerical error comparisons in FP16 and FP8 (e4m3).}
  \label{table:numerical_error}
  \begin{tabular}{|c|ccc|}
      \hline
      Method & Baseline FP16 & \faa FP16 & \fat FP16 \\
      RMSE & 3.2e-4 & \textbf{1.9e-4} & \textbf{1.9e-4} \\
      \hline
  \end{tabular}
  \vspace{1em}
  \begin{tabular}{|c|cccc|}
      \hline
      Method & Baseline FP8 & \fat FP8 & No block quant & No incoherent processing \\
      RMSE & 2.4e-2 & \textbf{9.1e-3} & 9.3e-3 & 2.4e-2 \\
      \hline
  \end{tabular}
\end{table}

\section{Dicussion, Limitations, Conclusion}
\label{sec:discussion}

With \fat, we have demonstrated that new programming techniques and hardware
features such as asynchrony and low-precision can have a dramatic impact on the
efficiency and accuracy of attention.
We are able to speed up attention by 1.5-2.0$\times$ times compared to \faa, and
reduce FP8 numerical error by 2.6$\times$ compared to standard per-tensor quantization.
Some limitations of our work that we hope to address in the future include: optimizing for
LLM inference, integrating a persistent kernel design into the FP8 kernel,\footnote{For our benchmarks, FP16 \fat has a persistent kernel and load balancing strategy, while FP8 \fat does not. This partly explains why FP8 \fat does not perform as well for small sequence length and causal masking compared to the FP8 cuDNN kernels.} and understanding the effects of low-precision attention in
large-scale training.
Though we have focused on Hopper GPUs in this work, we expect that the
techniques developed here will apply to other hardware accelerators.
We hope that a faster and more accurate primitive such as attention will unlock
new applications in long-context tasks.

\iftoggle{arxiv}{
\subsubsection*{Acknowledgments}

We are grateful to the NVIDIA CUTLASS team (especially Haicheng Wu, Aniket
Shivam, and Cris Cecka) for helping us understand Hopper's
programming model and for their library, which provides clean and powerful building blocks for the implementation of \fat.
We thank the cuDNN team for the idea of in-kernel transpose for FP8.
The idea of overlapping GEMMs and softmax was inspired by insightful
conversations with Christopher R{\'e}, Benjamin Spector, Aniket Shivam, and Markus Hoehnerbach.
The pingpong scheduling is adapted from the warp-specialized pingpong GEMM
implementation in CUTLASS.
We appreciate Driss Guessous for integrating \fa to PyTorch.
\fat has benefited from helpful discussions with Horace He on different attention
variants, with Hao Liu and Phil Wang on distributed attention, and with Daniel
Haziza and Chris De Sa on quantization.
We thank Meta, Together AI, and Princeton Language and Intelligence (PLI) for compute support.

}{}

\bibliography{ref}
\bibliographystyle{plainnat}

\newpage
\appendix
\section{Related Work}
\label{sec:related_work}

\paragraph{Attention variants and distributed attention}
Ever since attention became popular with the Transformer
architecture~\citep{vaswani2017attention}, there has been a large body of work
on approximating attention to scale it to longer sequences.
These approximation methods can generally be categorized into two classes:
sparse and low-rank.
Sparse attention only computes some entries of the attention matrix ($\mathrm{softmax}(\vQ
\vK^T)$) and assumes that other entries are zero.
Different methods have different ways of choosing which entries should be zero,
either with a fixed pattern~\citep{child2019generating}, with a sliding
window~\citep{beltagy2020longformer}, or with a dynamic pattern through
hashing~\citep{kitaev2020reformer} or routing~\citep{roy2020efficient}.
The low-rank approach instead assumes that the attention matrix has a low-rank
structure, and apply a pointwise nonlinearity to the query and
key~\citep{katharopoulos2020transformers} with random
projection~\citep{choromanski2021rethinking, peng2021random, xiong2021nystromformer}.
One can also combine the sparse and low-rank approximation for better
quality~\citep{zaheer2020bigbird,scatterbrain}.
However, these approximation methods typically do not offer the same model
quality as standard attention~\citep{tay2020efficient}, and so most large-scale
models do not employ these techniques.

There are other variants of attention aimed at reducing the size of the KV cache
to improve inference efficiency. Multi-query attention~\citep{shazeer2019fast} and grouped query
attention~\citep{ainslie2023gqa} tie different heads of $\vK$ and $\vV$, and
multiple query heads interact with the same key and value head.
Multi-head latent attention~\citep{deepseekv2} parameterizes the $\vK$ and $\vV$
as low-rank projections of a shared matrix to further reduce the KV cache size.
However, all of these approaches do not change the core computation
$\mathrm{softmax}(\vQ \vK^T) \vV$ during training and simply change how $\vQ, \vK, \vV$ are
obtained.
As a result, any efficiency or accuracy improvement to the standard attention
computation benefits these methods.

To extend to even longer context, attention computation can be distributed
across multiple GPUs.
Methods such as Ring attention~\citep{liu2023ring,liu2024world} and
variants~\citep{brandon2023striped} can reach a context length of up to 1
million.
They use \fa (or \faa) as a primitive, and so the improvement from \fat would
benefit these distributed attention methods as well.

\paragraph{Alternative architectures}
Motivated by the limitations of attention, a variety of alternative
architectures have been proposed.
They build on the connection between linear
attention~\citep{katharopoulos2020transformers} and recurrent neural networks
(RNNs).
RWKV~\citep{peng2023rwkv}, H3~\citep{dao2023hungry}, MEGA~\citep{ma2023mega},
Retnet~\citep{sun2023retentive}  enhance the expressivity of the simple
cumulative sum in linear attention with more sophisticated recurrences.
Mamba~\citep{gu2023mamba} and xLSTM~\citep{beck2024xlstm} use learnable
weighting for the recurrence and can match the quality of Transformers in
language modeling at small or medium scale.
These approaches can be connected to generalizations of linear attention through
the lens of the structure of the token-mixing matrix~\citep{dao2024transformers}.
These models have started to see some traction, seeing usage in some medium to
large-scale models such as Jamba~\citep{jamba}, Zamba~\citep{zamba},
Megalodon~\citep{ma2024megalodon}, and Mamba2-hybrid~\citep{waleffe2024empirical}.
For the highest quality, these SSM- and RNN-based models still employ
many layers of attention.
We expect that techniques to speed up attention presented in this work will be
useful to speedup these alternative architectures.

\paragraph{Low-precision attention}
Quantization is a promising approach to speed up attention, but they have mostly
focused on reducing the space for KV cache for inference efficiency.
QuIP~\citep{chee2024quip} and QuIP\#\citep{tseng2024quip} use incoherent processing to reduce the quantization,
and we adapted this technique for FP8 \fat.
Recent work suggests that for inference the KV cache is highly compressible down to 4-, 3-, or
even 2-bits~\citep{hooper2024kvquant, liu2024kivi}.
However, quantization during training is still challenging as higher precision
is typically required for stable training.

\paragraph{Hardware-aware Algorithms}
Our work presented in this paper focuses on the micro-architecture
specific tuning to leverage new instruction sets and adopt a natively
asynchronous programming model. There are other orthogonal axes for
hardware-aware algorithm co-design being explored.
A recent example of this is LeanAttention~\citep{sanovar2024-leanattention},
which recognizes the poor GPU occupancy and high memory bandwidth requirements
of the sequential token generation phase as primary bottlenecks for inference
and optimizes it via a smarter load balancing strategy similar to Stream-K
load balancing~\citep{streamk} to achieve nearly peak occupancy.
There is a large literature on optimizing GEMM for specific hardware that employs
many of the same techniques.
As an example, \citet{abdel2016batched} presents a high performance batched GEMM kernel on
K40c Graphics Processing Units (GPU) for both fixed and variable sizes,
proposing specialized GEMM designs
and a comprehensive autotuning process to deliver state-of-the-art 
performance.

\section{Addition Details on Algorithms}

\subsection{Asynchrony Through Warp Specialization for the Backward Pass}
\label{sec:algo_ws_bwd}

Similar to the forward pass~\cref{sec:algo_ws}, we use warp specialization to
handle asynchrony.
Instead of just a simple producer-consumer pattern in the forward pass, we add
one extra role of a $\vdQ$ writer, since we need to accumulate the value of $\vdQ$
produced by each thread block to the global value of $\vdQ$.
This $\vdQ$ accumulation introduces memory contention (many thread blocks writing to the same
location) so having a separate warp to handle this (along with asynchrony) will
avoid blocking the rest of the warps in the thread block to perform the next
computation (matmul).

We include the backward pass with warp specialization in~\cref{alg:flash3_wgmma_ws_bwd}.
\begin{algorithm}[H]
    \caption{\small\label{alg:flash3_wgmma_ws_bwd}\fat backward pass with warp specialization}
    \begin{algorithmic}[1]
\REQUIRE Matrices $\vQ, \vK, \vV, \vO, \vdO \in \mathbb{R}^{N \times d}$ in HBM,
logsumexp vector $L \in \mathbb{R}^N$ in HBM, block sizes $B_c$, $B_r$.
\STATE In a preprocessing kernel, compute $D = \mathrm{rowsum}(\vdO \circ \vO) \in \mathbb{R}^d$ (pointwise multiply), write
$D$ to HBM and divide it into $T_r$ blocks $D_1, \dots, D_{T_r}$ of size
$B_r$ each.
\STATE Divide $\vQ$ into $T_r = \left\lceil\frac{N}{B_r} \right\rceil$ blocks $\vQ_1, \dots, \vQ_{T_r}$ of size $B_r \times d$ each,
and divide $\vK, \vV$ in to $T_c = \left\lceil \frac{N}{B_c} \right\rceil$ blocks $\vK_1, \dots, \vK_{T_c}$ and
$\vV_1, \dots, \vV_{T_c}$, of size $B_c \times d$ each.
\STATE Divide $\vdO$ into $T_r$ blocks $\vdO_i, \dots, \vdO_{T_r}$
of size $B_r \times d$ each, and divide $L$ into $T_r$ blocks $L_i, \dots, L_{T_r}$ of size
$B_r$ each.
\STATE Initialize pipeline object to manage barrier synchronization with $s$-stage circular SMEM buffer.
\IF {in producer warpgroup}
\STATE Deallocate predetermined number of registers.
\STATE Issue load $\vK_j$ and $\vV_j$ from HBM to shared memory.
\STATE Upon completion, commit to notify consumer of the load of $\vK_j$ and $\vV_j$.
\FOR{$1 \le i \leq T_r$}
    \STATE Wait for the $(i\,\%\,s)$th stage of the buffer to be consumed.
    \STATE Issue loads of $\vQ_i, \vdO_i$ from HBM to shared memory at the $(i\,\%\,s)$th stage of the buffer.
    \STATE Upon completion, commit to notify consumers of the loads of $\vQ_i, \vdO_i$.
\ENDFOR
\ELSIF {in consumer warpgroups}
\STATE Reallocate predetermined number of registers as function of number of consumer warps.
\STATE On-chip, Initialize $\vdK_j = (0)_{B_c \times d}, \vdV_j = (0)_{B_c \times d}$ .
\STATE Wait for $\vK_j$ and $\vV_j$ to be loaded in shared memory.
\FOR{$1 \le i \leq T_r$}
\STATE Wait for $\vQ_i$ to be loaded in shared memory.
\STATE Load $L_i, D_i$ from HBM to on-chip SRAM.
\STATE On chip, compute $\vS_{i}^{(j)} = \vQ_i \vK_j^T \in \mathbb{R}^{B_r \times B_c}$
(SS-GEMM). Commit.
\STATE Wait for $\vdO_i$ to be loaded in shared memory.
\STATE On chip, compute $\vdP_{i}^{(j)} = \vdO_{i} \vV_j^\top \in \mathbb{R}^{B_r \times B_c}$
(SS-GEMM). Commit.
\STATE On chip, wait for $\vS_{i}^{(j)}$, then compute $\vP_{i}^{(j)} = \exp(\vS_{ij} - L_{i}) \in \mathbb{R}^{B_r \times B_c}$.
\STATE On chip, wait for $\vdP_i^{(j)}$, then compute $\vdS_{i}^{(j)} = \vP_{i}^{(j)} \circ (\vdP_{i}^{(j)} - D_i) \in \mathbb{R}^{B_r \times B_c}$.
\STATE On chip, compute
$\vdV_j \leftarrow \vdV_j + (\vP_{i}^{(j)})^\top \vdO_i \in \mathbb{R}^{B_c \times d}$ (RS-GEMM). Commit.
\STATE On chip, compute $\vdK_{j} \leftarrow \vdK_j + {\vdS_{i}^{(j)}}^\top \vQ_i \in \mathbb{R}^{B_c \times
  d}$ (RS-GEMM). Commit and wait for both $\vdV_j$ and $\vdK_j$.
\STATE On chip, compute $\vdQ_{i}^{(\mathrm{local})} = \vdS_{i}^{(j)} \vK_j \in
\mathbb{R}^{B_r \times d}$ (SS-GEMM), and write $\vdQ_i^{(\mathrm{local})}$ to smem. Notify
the $\vdQ$-writer.
\ENDFOR
\ELSIF{in $\vdQ$-writer warp}
\FOR{$1 \le i \leq T_r$}
\STATE Wait for $\vdQ_i^{(\mathrm{local})}$ to be ready in smem.
\STATE Using a semaphore, atomically add $\vdQ_i^{(\mathrm{local})}$ to $\vdQ_i$ in global memory.
\ENDFOR
\ENDIF
\end{algorithmic}
\end{algorithm}

\subsection{2-Stage Pipelining SASS Analysis}
\label{sec:2-stage-sass}

We give simplified SASS code for the inside of the consumer warpgroup mainloop.
\begin{small}
\begin{verbatim}
// Compute row_max
FMNMX.FTZ R0, R24, R6, !PT ;
SHFL.BFLY PT, R185, R2, 0x2, 0x1f ;
… FMNMX and SHFL.BFLY … 

// Apply exp2 and row_sum. Rescale O.
FMUL.FTZ R2, R4, UR9 ;
MUFU.EX2 R185, R184 ;                                            
FFMA.FTZ R24, R24, UR9, -R6.reuse ;
FADD.FTZ R24, R211, R24 ;                              
… FMUL, FFMA, FMUL, MUFU.EX2, FADD …

// FP32 -> FP16 conversion are interleaved with exp2, row_sum and O rescaling.
F2FP.F16.F32.PACK_AB R231, R25, R231 ;
… F2FP, FMUL, MUFU, FFMA, FADD ...

// Start the first WGMMA. Broken down into 8 HGMMAs.
// The first 7 HGMMAs are packed together.
WARPGROUP.ARRIVE ;
HGMMA.64x192x16.F32 R24, gdesc[UR44], RZ, !UPT ;
... HGMMA x 6 ...

// FP32->FP16, exp2, row_sum, O rescaling are interleaved with HGMMA. 
F2FP.F16.F32.PACK_AB R214, R214, R187 ; 
MUFU.EX2 R234, R5 ;
FADD.FTZ R237, R187, R2 ;
… F2FP, MUFU, FADD …

// The last HGMMA is issued here. No need to wait.
HGMMA.64x192x16.F32 R24, gdesc[UR44], R24, gsb0 ;

// Start the second WGMMA. Broken down into 12 HGMMAs.
// All 12 HGMMAs are packed together. Not interleaved with other instructions.
WARPGROUP.ARRIVE ;
HGMMA.64x128x16.F32 R120, R228, gdesc[UR8].tnspB, R120 ;
... HGMMA x 10 ...
HGMMA.64x128x16.F32 R120, R184, gdesc[UR8].tnspB, R120, gsb0 ;

// wgmma.wait_group at the end.
WARPGROUP.DEPBAR.LE gsb0, 0x0 ; 

\end{verbatim}
\end{small}

We make the following observations:
\begin{enumerate}
    \item Softmax is reordered to the very beginning, even before the first WGMMA. \TD{why?}
    \item The first WGMMA is interleaved with softmax and FP32 $\rightarrow$ FP16 datatype conversion of $\vS$. This indicates that WGMMA and non-WGMMAs are executed in parallel.
    \item \verb|exp2|, \verb|row\_sum|, O rescaling and FP32 $\rightarrow$ FP16 conversions are interleaved together. \TD{what's the benifit since these instructions are synced?} 
    \item The second WGMMA is not overlapped with other instructions, as expected.
\end{enumerate}
Overall, SASS shows that the 2-stage pipelining idea works as expected.

\subsection{3-Stage Pipelining Algorithm}
\label{sec:3-stage}
We experiment with a 3-stage pipelining algorithm to parallelize the first WGMMA from iteration $j+2$, softmax from iteration $j+1$, and the second WGMMA from iteration $j$. We describe this algorithm in \cref{alg:flash3_3_stage_wgmma}. 
This algorithm behaves worse than the 2-stage pipelining algorithm due to the reasons below:

\begin{figure}[ht]
    \centering
    \includegraphics[width=.95\linewidth]{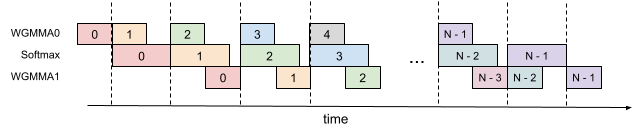}
    \caption{3-Stage Pipelining}
    \label{fig:3_stage_pipelining}
\end{figure}

\begin{algorithm}[H]
    \caption{\small\label{alg:flash3_3_stage_wgmma}\sysnameone 3-stage pipelining consumer warpgroup forward pass}
    \begin{algorithmic}[1]
      \REQUIRE Matrices $\vQ, \vK, \vV \in \mathbb{R}^{N \times d}$ in HBM, block sizes $B_c$, $B_r$.
      Each warpgroup reads 1 block Qi of size $B_r \times d$, $T_c = \left\lceil \frac{N}{B_c} \right\rceil$ blocks $\vK_1, \dots, \vK_{T_c}$ and
      $\vV_1, \dots, \vV_{T_c}$ of size $B_c \times d$.
      Each warpgroup writes 1 output block $\vO_i$ of size $B_r \times d$, and 1 logsumexp block $L_i$ of size $B_r$.
      \STATE Initialization. Load $\vQ_i$ from HBM to on-chip SRAM.
      Initialize $\vO_{i}, \ell_{i}, m_{i}, scale\_o$.
      \STATE Wait for the producer warpgroup loading $\vK_0$ from HBM to on-chip SRAM.
      \STATE Compute $\vS = \vQ_i \vK_0^T$ using WGMMA. Commit and wait.
      \STATE Compute $m_{i}$, $\tilde{\vP}_{i}$, $\ell_{i}$, $scale\_o$ based on $\vS$.
      \STATE Wait for the producer warpgroup loading $\vK_1$ from HBM to on-chip SRAM.
      \STATE Compute $\vS = \vQ_i \vK_1^T$ using WGMMA. Commit and wait.
      \FOR{$2 \le j < T_c - 2$}
        \STATE Wait for the producer warpgroup loading $\vK_j$ from HBM to on-chip SRAM.
        \STATE Compute $\vS\_next = \vQ_i \vK_{j}^T$ using WGMMA. Commit but do not wait.
        \STATE Wait for the producer warpgroup loading $\vV_{j-2}$ from HBM to on-chip SRAM.
        \STATE Rescale $\vO_{i}$ based on $scale\_o$.
        \STATE Compute
        $\vO_{i} = \vO_{i} + \tilde{\vP}_{i} \vV_{j-2}$ using WGMMA. Commit but do not wait.
        \STATE Compute $m_{i}$, $\tilde{\vP}_{i}\_next$, $\ell_{i}$, $scale\_o$ based on $\vS$.
        \STATE Wait for all previous WGMMAs.
        \STATE Copy $\vS\_next$ to $\vS$.
        \STATE Copy $\tilde{\vP}_{i}\_next$ to $\tilde{\vP}_{i}$.
      \ENDFOR
      \STATE Wait for the producer warpgroup loading $\vV_{T_c-2}$ from HBM to on-chip SRAM.
      \STATE Rescale $\vO_{i}$ based on $scale\_o$.
      \STATE Compute
      $\vO_{i} = \vO_{i} + \tilde{\vP}_{i} \vV_{T_c-2}$ using WGMMA. Commit and wait.
      \STATE Compute $m_{i}$, $\tilde{\vP}_{i}$, $\ell_{i}$, $scale\_o$ based on $\vS$.
      \STATE Wait for the producer warpgroup loading $\vV_{T_c-1}$ from HBM to on-chip SRAM.
      \STATE Rescale $\vO_{i}$ based on $scale\_o$.
      \STATE Compute
      $\vO_{i} = \vO_{i} + \tilde{\vP}_{i} \vV_{T_c-1}$ using WGMMA. Commit and wait.
      \STATE Epilogue. Rescale $\vO_{i}$ based on $\ell_{i}$. Compute $L_{i}$ based on $\ell_{i}$ and $m_{i}$. Write $\vO_{i}$ and $L_{i}$ to HBM as the $i$-th block of $\vO$ and $L$.
    \end{algorithmic}
  \end{algorithm}

\paragraph{Overlapping.} 
We expected that softmax can be overlapped with (the first WGMMA + the second WGMMA). However, the compiler doesn't cooperate in this way.
SASS code shows that only the first WGMMA is overlapped with softmax, while the second WGMMA is not. It's not clear why the compiler chooses to reorder instructions in this way.

\paragraph{Register pressure.}
This algorithm requires more registers compared to the 2-stage pipelining algorithm. 
In theory, it needs to store an extra $\tilde{\vP}_{i}$ and $scale\_o$, which is of size $B_r \times B_c \times \text{sizeof}(\text{input\_data\_type}) + B_r \times \text{sizeof}(\text{float})$.
As a result, a smaller block size needs to be chosen.

\section{Addition Details on Experiments and Benchmarking}

\subsection{System and libraries}
\label{sec:system}

We benchmark the speed on an H100 80GB SXM5 (700W).
We generally use the latest versions of the libraries, at the time of writing
(May 2024).
Specifically, we use:
\begin{itemize}
\item CUDA 12.3
\item cuDNN 9.1.1.17
\item CUTLASS 3.5
\item \fa 2.5.8
\item Triton nightly 3.0.0.post20240424212437
\item PyTorch 2.3.0
\end{itemize}

To reduce variability, we fix the GPU clock speed to 1830MHz (clock speed used
to calculate the 989 TFLOPS FP16 theoretical max throughput).
We repeat the benchmarks 100 times and take the average timing.

\subsection{FP8 Attention Full Results}
\label{sec:benchmark_attn_fp8_full}

We use following sequence lengths: 512, 1024, 2048, 4224, 8448, 16896.
When sequence length $\geq$ 4k, we make it also divisible by 132 (number of SMs in H100 SXM5) to avoid wave quantization.

\begin{figure}[ht]
  \centering
  \begin{subfigure}{.5\textwidth}
    \centering
    \includegraphics[width=.95\linewidth]{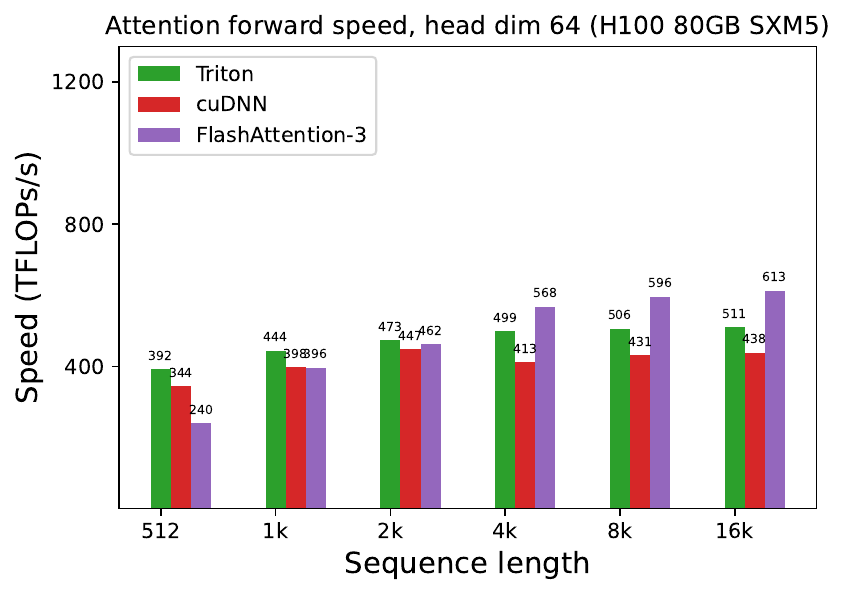}
    \caption{Forward, without causal mask, head dim 64}
  \end{subfigure}%
  \begin{subfigure}{.5\textwidth}
    \centering
    \includegraphics[width=.95\linewidth]{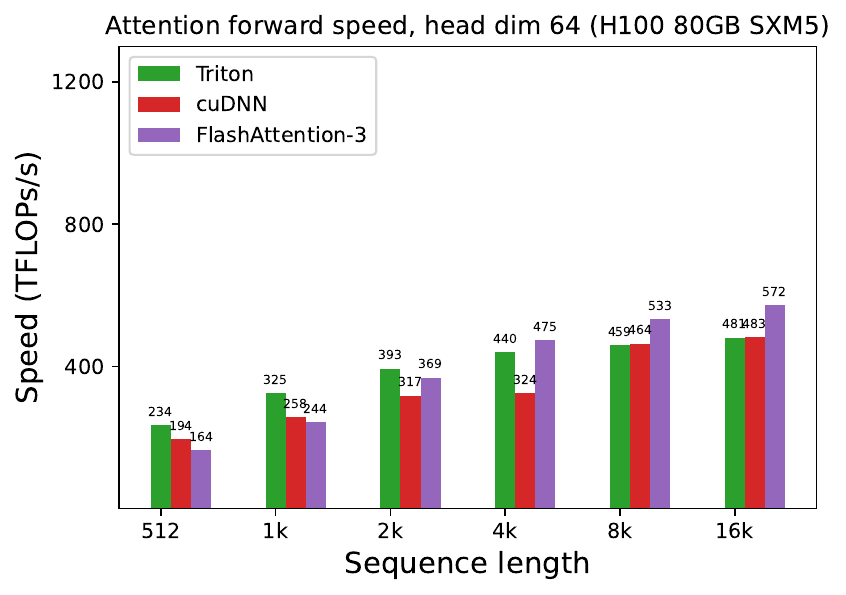}
    \caption{Forward, with causal mask, head dim 64}
  \end{subfigure}
  \begin{subfigure}{.5\textwidth}
    \centering
    \includegraphics[width=.95\linewidth]{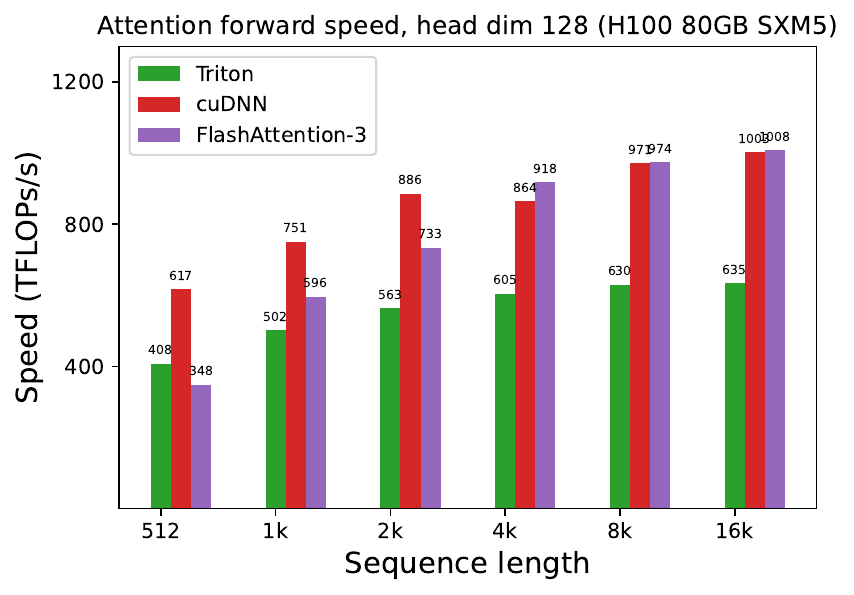}
    \caption{Forward, without causal mask, head dim 128}
  \end{subfigure}%
  \begin{subfigure}{.5\textwidth}
    \centering
    \includegraphics[width=.95\linewidth]{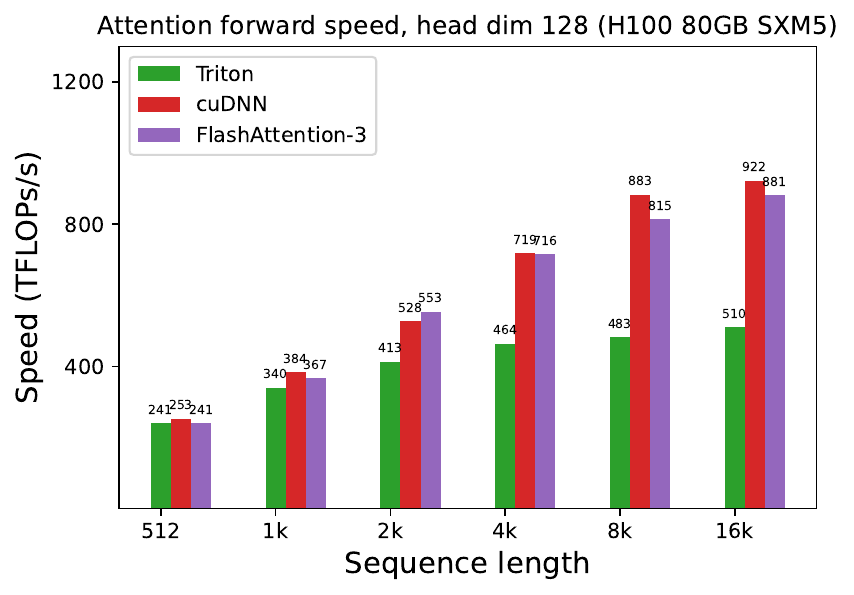}
    \caption{Forward, with causal mask, head dim 128}
  \end{subfigure}
  \begin{subfigure}{.5\textwidth}
    \centering
    \includegraphics[width=.95\linewidth]{figs/flash3_h100_fp8_causal_False_hdim_256_fwd_speed.pdf}
    \caption{Forward, without causal mask, head dim 256}
  \end{subfigure}%
  \begin{subfigure}{.5\textwidth}
    \centering
    \includegraphics[width=.95\linewidth]{figs/flash3_h100_fp8_causal_True_hdim_256_fwd_speed.pdf}
    \caption{Forward, with causal mask, head dim 256}
  \end{subfigure}
  \caption{Attention forward speed (FP8) on H100 GPU}
  \label{fig:benchmark_attn_fwd_full}
\end{figure}

\iftoggle{arxiv}{}{
\newpage
\section*{NeurIPS Paper Checklist}

The checklist is designed to encourage best practices for responsible machine learning research, addressing issues of reproducibility, transparency, research ethics, and societal impact. Do not remove the checklist: {\bf The papers not including the checklist will be desk rejected.} The checklist should follow the references and follow the (optional) supplemental material.  The checklist does NOT count towards the page
limit.

Please read the checklist guidelines carefully for information on how to answer these questions. For each question in the checklist:
\begin{itemize}
    \item You should answer \answerYes{}, \answerNo{}, or \answerNA{}.
    \item \answerNA{} means either that the question is Not Applicable for that particular paper or the relevant information is Not Available.
    \item Please provide a short (1–2 sentence) justification right after your answer (even for NA).
\end{itemize}

{\bf The checklist answers are an integral part of your paper submission.} They are visible to the reviewers, area chairs, senior area chairs, and ethics reviewers. You will be asked to also include it (after eventual revisions) with the final version of your paper, and its final version will be published with the paper.

The reviewers of your paper will be asked to use the checklist as one of the factors in their evaluation. While "\answerYes{}" is generally preferable to "\answerNo{}", it is perfectly acceptable to answer "\answerNo{}" provided a proper justification is given (e.g., "error bars are not reported because it would be too computationally expensive" or "we were unable to find the license for the dataset we used"). In general, answering "\answerNo{}" or "\answerNA{}" is not grounds for rejection. While the questions are phrased in a binary way, we acknowledge that the true answer is often more nuanced, so please just use your best judgment and write a justification to elaborate. All supporting evidence can appear either in the main paper or the supplemental material, provided in appendix. If you answer \answerYes{} to a question, in the justification please point to the section(s) where related material for the question can be found.

IMPORTANT, please:
\begin{itemize}
    \item {\bf Delete this instruction block, but keep the section heading ``NeurIPS paper checklist"},
    \item  {\bf Keep the checklist subsection headings, questions/answers and guidelines below.}
    \item {\bf Do not modify the questions and only use the provided macros for your answers}.
\end{itemize}

\begin{enumerate}

\item {\bf Claims}
    \item[] Question: Do the main claims made in the abstract and introduction accurately reflect the paper's contributions and scope?
    \item[] Answer: \answerYes{} %
    \item[] Justification: Abstract and intro reflects the paper's contribution.

\item {\bf Limitations}
    \item[] Question: Does the paper discuss the limitations of the work performed by the authors?
    \item[] Answer: \answerYes{} %
    \item[] Justification: Discussed in~\cref{sec:discussion}

\item {\bf Theory Assumptions and Proofs}
    \item[] Question: For each theoretical result, does the paper provide the full set of assumptions and a complete (and correct) proof?
    \item[] Answer: \answerNA{} %
    \item[] Justification: The paper does not include theoretical results.

    \item {\bf Experimental Result Reproducibility}
    \item[] Question: Does the paper fully disclose all the information needed to reproduce the main experimental results of the paper to the extent that it affects the main claims and/or conclusions of the paper (regardless of whether the code and data are provided or not)?
    \item[] Answer: \answerYes{} %
    \item[] Justification: Detailed information in~\cref{sec:system}.

\item {\bf Open access to data and code}
    \item[] Question: Does the paper provide open access to the data and code, with sufficient instructions to faithfully reproduce the main experimental results, as described in supplemental material?
    \item[] Answer: \answerNo{} %
    \item[] Justification: The code will be released with a permissive license in the near future.

\item {\bf Experimental Setting/Details}
    \item[] Question: Does the paper specify all the training and test details (e.g., data splits, hyperparameters, how they were chosen, type of optimizer, etc.) necessary to understand the results?
    \item[] Answer: \answerNA{} %
    \item[] Justification: The paper does not include training models.

\item {\bf Experiment Statistical Significance}
    \item[] Question: Does the paper report error bars suitably and correctly defined or other appropriate information about the statistical significance of the experiments?
    \item[] Answer: \answerNo{} %
    \item[] Justification: Not necessary for speed benchmarks since we already
      take average of a large number (30) of trials.

\item {\bf Experiments Compute Resources}
    \item[] Question: For each experiment, does the paper provide sufficient information on the computer resources (type of compute workers, memory, time of execution) needed to reproduce the experiments?
    \item[] Answer: \answerYes{} %
    \item[] Justification: In~\cref{sec:system}

\item {\bf Code Of Ethics}
    \item[] Question: Does the research conducted in the paper conform, in every respect, with the NeurIPS Code of Ethics \url{https://neurips.cc/public/EthicsGuidelines}?
    \item[] Answer: \answerYes{} %
    \item[] Justification: Yes

\item {\bf Broader Impacts}
    \item[] Question: Does the paper discuss both potential positive societal impacts and negative societal impacts of the work performed?
    \item[] Answer: \answerNo{} %
    \item[] Justification: The paper focuses on foundational research and not
      tied to a particular application.

\item {\bf Safeguards}
    \item[] Question: Does the paper describe safeguards that have been put in place for responsible release of data or models that have a high risk for misuse (e.g., pretrained language models, image generators, or scraped datasets)?
    \item[] Answer: \answerNA{} %
    \item[] Justification: No release of high-risk data or models.

\item {\bf Licenses for existing assets}
    \item[] Question: Are the creators or original owners of assets (e.g., code, data, models), used in the paper, properly credited and are the license and terms of use explicitly mentioned and properly respected?
    \item[] Answer: \answerYes{} %
    \item[] Justification: Yes

\item {\bf New Assets}
    \item[] Question: Are new assets introduced in the paper well documented and is the documentation provided alongside the assets?
    \item[] Answer: \answerNA{} %
    \item[] Justification: No new assets.

\item {\bf Crowdsourcing and Research with Human Subjects}
    \item[] Question: For crowdsourcing experiments and research with human subjects, does the paper include the full text of instructions given to participants and screenshots, if applicable, as well as details about compensation (if any)?
    \item[] Answer: \answerNA{} %
    \item[] Justification: No crowdsourcing or human subjects.

\item {\bf Institutional Review Board (IRB) Approvals or Equivalent for Research with Human Subjects}
    \item[] Question: Does the paper describe potential risks incurred by study participants, whether such risks were disclosed to the subjects, and whether Institutional Review Board (IRB) approvals (or an equivalent approval/review based on the requirements of your country or institution) were obtained?
    \item[] Answer: \answerNA{} %
    \item[] Justification: No crowdsourcing or human subjects.

\end{enumerate}

}

\end{document}